\documentclass[11pt]{article}

\usepackage[preprint]{acl}

\usepackage{times}
\usepackage{latexsym}

\usepackage[T1]{fontenc}

\usepackage[utf8]{inputenc}

\usepackage{microtype}

\usepackage{inconsolata}

\usepackage{graphicx}

\usepackage{amsmath}
\usepackage{mathtools}
\usepackage{amsthm}
\usepackage{amsfonts}
\usepackage{amssymb}

\usepackage[utf8]{inputenc}
\usepackage[T1]{fontenc}

\usepackage{url}
\usepackage{nicefrac}
\usepackage{xcolor}
\usepackage{enumitem}
\usepackage{xspace}
\usepackage{etoolbox}

\usepackage{titletoc}
\usepackage{multicol}
\usepackage{multirow}
\usepackage{colortbl}
\usepackage{tabularx}
\usepackage{makecell}
\usepackage{caption}
\usepackage{booktabs}
\usepackage[most]{tcolorbox}
\usepackage{wrapfig}
\usepackage{rotating}

\usepackage{pifont}
\usepackage{bbding}

\usepackage{listings}
\usepackage{fvextra}

\usepackage{amsmath,amsfonts,bm}

\def\eqref#1{equation~\ref{#1}}

\def\1{\bm{1}}

\DeclareMathAlphabet{\mathsfit}{\encodingdefault}{\sfdefault}{m}{sl}
\SetMathAlphabet{\mathsfit}{bold}{\encodingdefault}{\sfdefault}{bx}{n}

\definecolor{myblue}{rgb}{0.2, 0.3, 0.6}
\newcommand{\modelname}{\textsc{SupraBench}\xspace}
\newcommand{\corpusname}{\textsc{SupraPmc}\xspace}

\title{\modelname: A Benchmark for Supramolecular Chemistry}

\author{
    \textbf{Tianyi Ma}$^{1,*}$\;
    \textbf{Yijun Ma}$^{1,*}$\;
    \textbf{Zehong Wang}$^{1,\dagger}$\;
    \textbf{Weixiang Sun}$^{1}$\;
    \textbf{Ziming Li}$^{2}$\;
    \\
    \textbf{Connor R. Schmidt}$^{1}$\;
    \textbf{Chuxu Zhang}$^{2}$\; 
    \textbf{Matthew J. Webber}$^{1}$\;
    \textbf{Yanfang Ye}$^{1,\dagger}$
    \\
    \textsuperscript{1} University of Notre Dame \quad 
    \textsuperscript{2} University of Connecticut
    \\
    $^*$ Equal Contribution \quad
    $^\dagger$ Corresponding Authors
    \\
    \texttt{<tma2,yma7,zwang43,yye7>@nd.edu}
}

\begin{document}
\maketitle
\begin{abstract}
    Supramolecular chemistry, which includes the study of non-covalent host–guest assemblies, has shown advances in various applications.
However, designing host–guest systems remains a time-consuming process, requiring days of dry-lab verification per candidate pair.
Although LLMs have emerged as a fast alternative with strong performance on molecular binding tasks, no benchmark currently systematically evaluates LLMs for host–guest reasoning across fundamental supramolecular chemistry tasks, \textit{e.g.}, binding affinity prediction.
To this end, we collaborate with domain experts to release the first \textbf{Supra}molecular \textbf{Bench}mark, called \textbf{\modelname}, to evaluate LLMs in chemistry reasoning.
Specifically, we design four fundamental tasks, \textit{i.e.}, binding affinity prediction, top-binder selection, solvent identification, and host–guest description, plus an auxiliary vision-based task for molecular identification.
We also release \textbf{\corpusname}, a curated 16M-token corpus of \textbf{Supra}molecular chemistry articles distilled from Europe \textbf{PMC}, to support the adaptation to the supramolecular domain.
We benchmark a broad range of open and proprietary LLMs and find that LLMs leave substantial headroom across all tasks.
Domain adaptation pretraining over \corpusname transfers cleanly to in-distribution regression but trades off against strict letter-format output.
Moreover, the difficulty profile differs sharply across task families, revealing distinct failure modes that indicate specific gaps in current supramolecular chemical reasoning.
Our source codes and benchmark datasets are available at \href{https://github.com/Tianyi-Billy-Ma/SupraBench}{here}.

\end{abstract}

\section{Introduction}
\label{sec:intro}

\begin{figure}[t]
    \centering
    \includegraphics[width=0.95\linewidth]{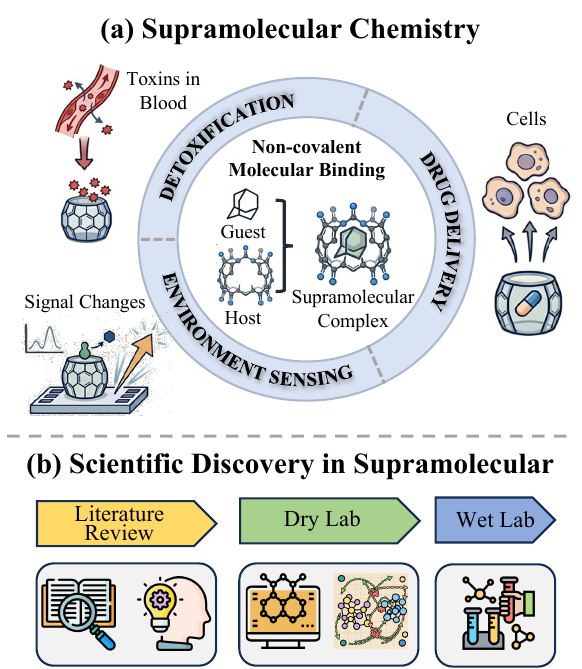}
    \caption{\textbf{Overview of host–guest supramolecular chemistry.} (a) Studies and applications in supramolecular chemistry. (b) Classic scientific discovery pipeline in supramolecular chemistry.}
    \label{fig:intro}
\end{figure}

Supramolecular chemistry includes the study of reversible non-covalent host–guest assemblies and represents a third bond-design paradigm alongside ionic and covalent chemistry, with pioneering work recognized by the 1987 Nobel Prize in Chemistry \citep{lehn1995supramolecular}.
Its applications are already widespread, including drug delivery~\citep{loftsson2010cyclodextrins,webber2017drug}, chemical sensing~\citep{you2015sensing,kolesnichenko2017practical}, and in vivo detoxification of pharmaceutical and anesthetic agents~\citep{brockett2023pillar6maxq,deng2020sequestration}, as illustrated in Figure~\ref{fig:intro}(a).
For example, sugammadex~\citep{bom2002sugammadex}, a $\gamma$-cyclodextrin host that selectively encapsulates the muscle relaxant rocuronium, has been approved in 75 countries and has become a standard reversal agent in operating rooms worldwide, reducing postoperative pulmonary complications relative to prior protocols~\citep{olesnicky2024effect}.

Although these applications reach patients worldwide to reduce morbidity at scale, each success takes decades of expert iteration to deliver, and designing such a host–guest system remains time-consuming.
Specifically, when using computational tools to design candidate host scaffolds and guests that match a target binding profile~\citep{thordarson2011determining}, each short-listed pair may be verified computationally through classical simulation approaches, \textit{e.g.}, density functional theory (DFT) or molecular dynamics (MD), to estimate the binding free energy before committing to wet-lab chemical synthesis~\citep{yin2017overview, mobley2017predicting}. 
Each computational verification may require days of computation, even on modern High-Performance Computing clusters~\citep{yin2017overview}.
Over the entire design pipeline, this read-then-simulate loop dominates the calendar long before any molecule reaches the bench.

With recent advances in large language models (LLMs) \cite{ye2025llms4all,chen2025obvious,chen2025clear, ma2025llm, ma2026non}, existing studies have attempted to leverage these as a fast alternative to the time-consuming DFT or MD pipelines at the hypothesis-generation stage in molecular binding prediction tasks.
Across binding-related tasks, structure–prediction foundation models such as Boltz~\citep{passaro2025boltz} and follow-ups~\citep{feng2025foundation} now approach FEP-level affinity, LLM-based methods predict drug-target interactions~\citep{li2025druglm,ye2026latentchem}, and zero-shot kinase-inhibitor binding at accuracies that exceed classical docking~\citep{liu2024gpt4kinase}.
In particular, \citet{parrilla2024electron} proposes an electron-density-conditioned GPT that generates candidate host–guest binders for CB[n] and a family of metal–organic cages, and experimentally validates several previously unreported guests.
Despite these advances, no benchmark currently evaluates whether modern LLMs can support host–guest reasoning tasks.

To this end, we introduce the first \textbf{Bench}mark for \textbf{Supra}molecular chemistry, called \textbf{\modelname}, to evaluate LLMs in host–guest reasoning tasks. 
Specifically, we collaborate with experts in supramolecular chemistry to carefully define four fundamental tasks, \textit{i.e.}, binding affinity prediction, top-binder selection, solvent identification, and host–guest description, along with one auxiliary vision-based task for molecular identification.
Moreover, we release \textbf{\corpusname}, a curated 16M-token corpus of \textbf{Supra}molecular chemistry articles obtained from Europe~\textbf{PMC}~\citep{europe2015europe}, to support future research in domain adaptation.
Across the five tasks, we benchmark a broad range of open and proprietary LLMs and find that general-purpose models leave substantial headroom on every task, that domain adaptation closes a measurable but uneven portion of this gap, and that the difficulty profile differs sharply across task families, exposing distinct failure modes.
We believe our benchmark \modelname, along with our released supramolecular chemistry text corpus \corpusname, can contribute to the relevant research communities.
Our main contributions are summarized as follows:
\begin{itemize}[leftmargin=*,nosep]
    \item \textbf{Supramolecular Benchmark.}
    We introduce \modelname, the first supramolecular benchmark comprising four fundamental tasks and an auxiliary vision-based task, to evaluate LLMs under a unified protocol and a single set of metrics.
    \item \textbf{Supramolecular Text Corpus.}
    We release \corpusname, a 16M-token corpus of supramolecular chemistry articles, that can contribute to the research communities in this field.
    \item \textbf{Systematic Evaluation and Insights.}
    We evaluate a representative set of open and proprietary-based LLMs, including a domain-adapted variant trained over \corpusname. 
    Among the evaluation results, we reveal actionable insights into the strengths and limitations of existing LLMs for supramolecular chemistry tasks.
\end{itemize}

\begin{figure*}[t]
    \centering
    \includegraphics[width=0.98\linewidth]{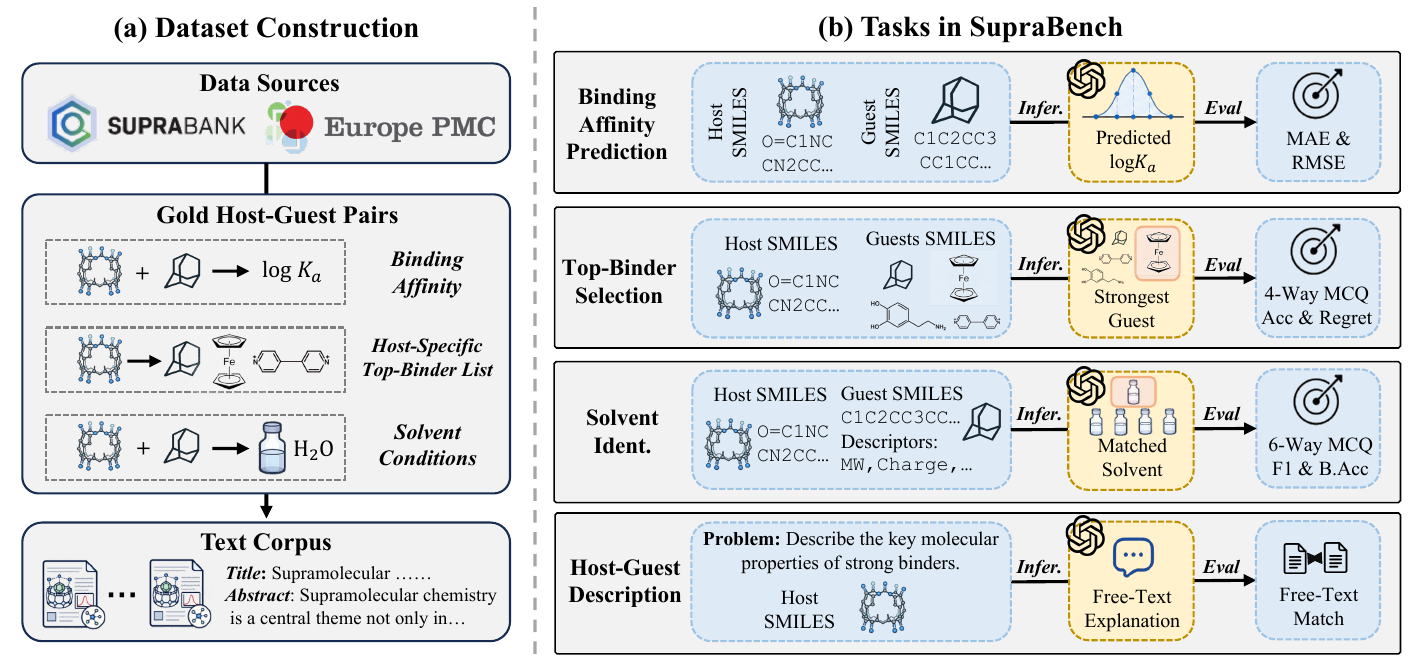}
    \caption{\textbf{Overview of \modelname.} (a) We obtain gold host–guest interaction records from SupraBank, and text corpus of supramolecular chemistry articles from Europe PMC. (b) \modelname contains four fundamental tasks in supramolecular chemistry reasoning, including binding affinity prediction, top-binder selection, solvent identification, and host–guest description. }
    \label{fig:overview}
\end{figure*}

\section{Background}
\label{sec:background}
We discuss the key relevant works in this section, and leave detailed related works in Appendix~\ref{app:related_works}.

\paragraph{LLMs for Chemistry.}
LLMs have advanced chemistry along two complementary directions.
The first line of work~\citep{zhang2024chemllm, zhao2025chemdfm, yu2024llasmol, fang2024molinstructions} develops {domain-adapted LLMs} that trains on chemistry-specific corpora and instruction data.
Moreover, several studies introduce chemistry foundation models, such as Galactica~\citep{taylor2022galactica}, MolT5~\citep{edwards2022molt5}, and nach0~\citep{livne2024nach0}.
The second direction introduces LLM-based agents, equipped with external tools for laboratory automation, \textit{e.g.}, ChemCrow~\citep{bran2024chemcrow}, Coscientist~\citep{boiko2023coscientist}, and ChemAgent~\citep{tang2025chemagent}.
Across both directions, evaluation has primarily focused on small molecule tasks such as property prediction, retrosynthesis, and reaction yield estimation, leaving the supramolecular host–guest setting that \modelname targets largely unaddressed.

\paragraph{Chemistry Benchmarks.}
Chemistry benchmarks have evolved alongside model capabilities.
Early suites such as MoleculeNet~\citep{wu2018moleculenet} have standardized molecular property prediction, while GuacaMol~\citep{brown2019guacamol} and MOSES~\citep{polykovskiy2020molecular} provided canonical evaluations for generative chemistry.
LLM-specific benchmarks~\citep{guo2023chemllmbench, mirza2025chembench} scale a similar idea to LLM evaluation, and broader scientific benchmarks~\citep{wang2024scibench, sun2024scieval, laurent2024labbench} evaluate quantitative and research assistant capabilities.
Every existing LLM-targeted benchmark mainly focuses on single-molecule or single-system reasoning.
The closest molecular-modeling analog is the SAMPL host–guest blind challenge~\citep{amezcua2022sampl8}, which calibrates physics-based free-energy methods on a handful of curated pairs rather than evaluating LLMs.
\modelname fills this gap by evaluating LLMs on supramolecular host–guest reasoning under a unified protocol.

\section{\textsc{\modelname}}
\label{sec:benchmark}

\begin{table}[t]
\centering
\small
\setlength{\tabcolsep}{4pt}
\renewcommand{\arraystretch}{1.05}
\begin{tabular}{cc}
\textbf{(a) \# Sample per task} & \textbf{(b) \# Sample in top-4 hosts} \\[2pt]
\begin{tabular}{l@{\hspace{3em}}c}
    \toprule
    \textbf{Task} & \textbf{\# Samples} \\
    \midrule
    \textbf{BAP} & $2{,}609$ \\
    \textbf{TBS} & $2{,}264$ \\
    \textbf{SID} & $2{,}172$ \\
    \textbf{HGD} & $135$ \\
    \bottomrule
\end{tabular}
&
\begin{tabular}{lrrr}
    \toprule
    \textbf{Host} & \textbf{BAP} & \textbf{TBS} & \textbf{SID} \\
    \midrule
    CB[8]      & $261$ & $200$ & $571$ \\
    CB[7]      & $217$ & $200$ & $217$ \\
    $\beta$-CD & $201$ & $200$ & $264$ \\
    $p$-SC4    & $144$ & $144$ & $225$ \\
    \bottomrule
\end{tabular}
\\
\end{tabular}
\caption{\textbf{Dataset statistics for \modelname.} Here, BAP, TBS, SID, and HGD denote binding affinity prediction, top-binder selection, solvent identification, and host–guest description tasks, respectively.}
\label{tab:stats}
\end{table}

\subsection{Dataset Construction}
\label{sec:benchmark:curation}
\paragraph{Anchor Records.} 
We obtain the anchor host–guest binding records from SupraBank~\citep{suprabank}, a public repository that curates the experimentally reported supramolecular interaction records. 
Specifically, we leverage AutoData~\cite{ma2026autodata} to crawl the host–guest binding measurements associated with the molecular metadata, including host and guest names, identifiers, images, canonical SMILES strings, binding constants, and solvent conditions. 
Afterward, we obtain $5{,}362$ raw binding records over $2{,}466$ components.
\paragraph{Post-Processing.}
The collected raw records remain noisy: molecular identifiers are inconsistent across publications, binding constants are reported under heterogeneous experimental conditions (\textit{e.g.}, solvent, temperature, pH), and the same host–guest pair is frequently measured multiple times with occasional outliers.
To address these issues, we employ a six-step cleaning pipeline (detailed in Appendix~\ref{app:dataset}), \textit{i.e.}, numeric parsing, organic-solvent filtering, default-condition imputation, van't Hoff temperature correction, per-pair averaging, and outlier removal.
As a result, we obtain $4{,}635$ high-quality records across $2{,}008$ unique components.

\begin{figure}[t]
\centering
\resizebox{0.95\linewidth}{!}{
\begin{tcolorbox}[
    colback=gray!5,
    colframe=black!70,
    title=\textbf{\modelname Record-444913-637563},
    fonttitle=\bfseries,
    boxsep=3pt, left=5pt, right=5pt, top=3pt, bottom=3pt,
]
    \small
    \begin{tabular}{@{}l l@{}}
            \multicolumn{2}{@{}l@{}}{\emph{Host Identifiers}} \\
            \quad Name        & $\alpha$-cyclodextrin \\
            \quad PubChem CID & 444913 \\
            \quad SMILES      & \texttt{OC[C@H]1O[C@@H]2O...} \\
            \midrule
            \multicolumn{2}{@{}l@{}}{\emph{Guest Identifiers}} \\
            \quad Name        & \emph{trans}-anethole \\
            \quad PubChem CID & 637563 \\
            \quad SMILES      & \texttt{C/C=C/C1=CC=C(OC)C=C1} \\
            \midrule
            \multicolumn{2}{@{}l@{}}{\emph{Measurement}} \\
            \quad Binding constant $K_a$ & $710~\mathrm{M^{-1}}$ \\
            \quad $\log K_a$             & $2.85$ \\
            \quad Binding free energy $\Delta G$ & $-16.27~\mathrm{kJ/mol}$ \\
            \quad pH                     & $7.0$ \\
            \quad Temperature            & $298~\mathrm{K}$ \\
    \end{tabular}
    \vspace{1mm}
    \hrule
    \vspace{1mm}
    \noindent\small\emph{Images}
    \par
    \vspace{1mm}
    \begin{minipage}[t]{0.49\linewidth}
        \centering
        \includegraphics[width=0.95\linewidth]{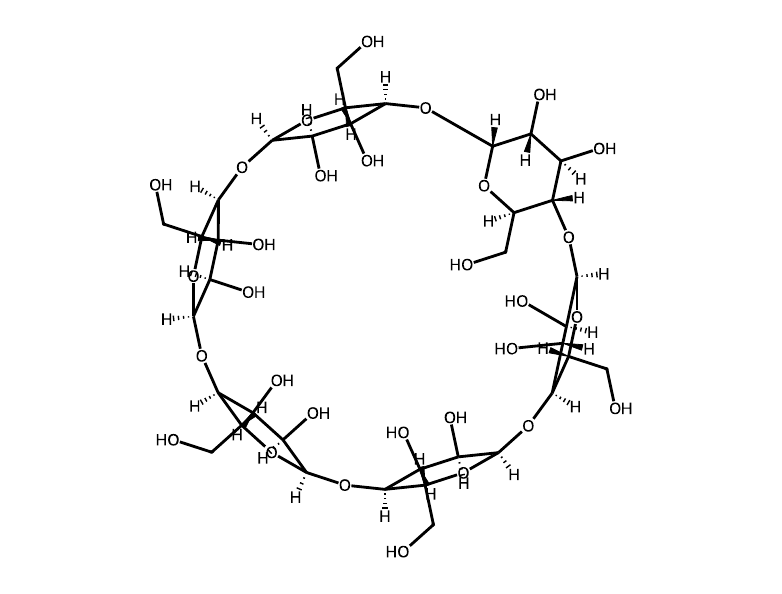}
        \par\small \textbf{Host:} $\alpha$-cyclodextrin
    \end{minipage}\hfill
    \begin{minipage}[t]{0.49\linewidth}
        \centering
        \includegraphics[width=0.95\linewidth]{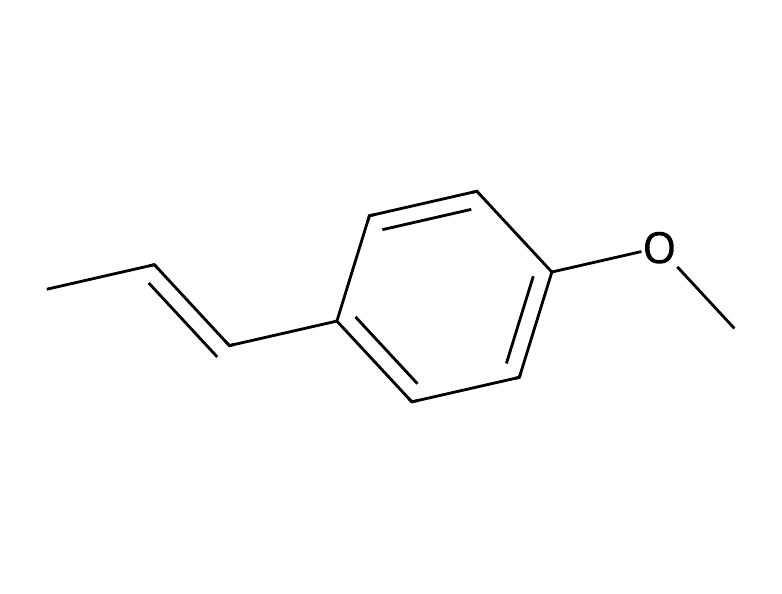}
        \par\small \textbf{Guest:} \emph{trans}-anethole
    \end{minipage}
\end{tcolorbox}
}
\caption{\textbf{Binding record example in \modelname}}
\label{fig:example_record}
\end{figure}

\subsection{Task Construction}
\label{sec:benchmark:tasks}

\modelname comprises four fundamental tasks, each targeting a distinct supramolecular reasoning capability, plus an auxiliary multimodal task that probes whether the model can identify a molecule from its 2D structural drawing.
The data statistics are provided in Table~\ref{tab:stats}.

\paragraph{Binding Affinity Prediction (BAP).}
This task evaluates whether a model can predict the affinity of a host–guest interaction directly from molecular structure, the central design quantity across supramolecular applications, ranging from cyclodextrin-based drug formulations~\citep{loftsson2010cyclodextrins} to cucurbituril-based solubility enhancers~\citep{webber2017drug}.
We formulate this task as a regression problem in which the model predicts $\log K_a$ under standard aqueous conditions, given the host and guest SMILES strings.

\paragraph{Top-Binder Selection (TBS).}
This task challenges a model to reason about relative binding selectivity, \textit{i.e.}, distinguishing the strongest binder among chemically similar candidates, a capability that underpins binder screening and sensor design~\citep{you2015sensing,kolesnichenko2017practical}.
We formulate this task as a multiple-choice question, in which the model picks the strongest binder among four candidate guests for a host.

\paragraph{Solvent Identification (SID).}
This task tests whether a model can infer the experimental measurement context from molecular structure alone, since binding constants are only comparable within a shared solvent regime~\citep{kolesnichenko2017practical}.
We formulate this task as a six-way classification wherein the model identifies in which of water, DMSO, MeCN, MeOH, CHCl$_3$, or CH$_2$Cl$_2$ the reported binding constant is measured.

\paragraph{Host–Guest Description (HGD).}
This task examines whether a model has physicochemical knowledge of host–guest pairs, a capability that underpins inverse-design workflows such as the macrocyclic sequestrant search that produced sugammadex~\citep{bom2002sugammadex} and pillar[6]uril opioid antidotes~\citep{brockett2023pillar6maxq}.
We formulate this task as an open-ended host–guest QA task with two complementary subtypes: the \emph{reverse} subtype asks the model to describe candidate hosts and their structural features for a given guest, while the \emph{forward} subtype asks the model to describe the property profile and representative examples of guests that bind a given host.

\paragraph{Molecular Identification (MI).}
This auxiliary task examines whether a multimodal model can recover a molecule's identity from its 2D structural drawing, a vision-grounded capability that complements the four text-based tasks above and exposes precise bond-level reasoning.
We formulate it as image-to-SMILES generation in which the model receives a single rendered structure image of a host or guest molecule and emits the corresponding canonical SMILES string. Further task construction details are provided in Appendix~\ref{app:task}.

\subsection{\corpusname Construction}
To enable the adaptation for the general-purpose LLMs, we further release \textbf{\corpusname}, a 16M-token \textbf{Supra}molecular chemistry text corpus obtained from Europe~\textbf{PMC}~\citep{europe2015europe}, comprising existing articles relevant to the field.

\paragraph{Anchor Corpus.}
Our text corpus for supramolecular chemistry is obtained from Europe PMC, an open-access repository that indexes biomedical and life sciences citations, via its REST endpoint.
We initially retrieve the full abstract index spanning the publication years $1900$ through $2026$, leading to over forty million articles, of which over eight million contain the full article in XML format.

\paragraph{Supramolecular Text Corpus.}
With the anchor text corpus, we further filter the supramolecular text corpus by issuing $19$ topical queries that span the supramolecular principle sub-areas, \textit{e.g.}, host–guest chemistry, self-assembly, and molecular recognition, \textit{etc.}
As a result, we obtain $420{,}950$ raw filtered articles. 
Note that these raw, filtered articles still contain residual biomedical contamination, such as papers that mention \emph{host cells} rather than \emph{host molecules}.
To this end, we further refine the result with a transparent rule-based filter that combines a positive bank of supramolecular keywords with two reject banks of biomedical contamination.
Afterward, we obtain a high-precision filtered split of $133{,}867$ articles. 
More details about the filtering logic and keyword lists are discussed in Appendix~\ref{app:text_corpus}.

\subsection{Evaluation Protocol}
\label{sec:benchmark:protocol}
\paragraph{Prompt Strategies.} To comprehensively evaluate the model performance, we implement three prompting strategies:
(i) \textbf{Base}, a zero-shot direct prompt that issues the question and answer schema with no exemplars.
(ii) \textbf{Few-Shot}, which prepends three to five in-domain demonstrations sampled from the pool of the same task.
Note that these few-shot examples are not evaluated as a test set.
and 
(iii) \textbf{CoT}~\citep{wei2022chain}, which requires an explicit reasoning before the final answer.

\paragraph{Evaluation Metrics.}
We employ widely used evaluation metrics for tasks in \modelname.
For binding affinity prediction, we report mean absolute error (MAE) and root mean squared error (RMSE) over extracted $\log K_a$ values.
For top-binder selection, we report letter accuracy and host–guest regret over the four candidate guests.
For solvent identification, we report both class-balanced accuracy and Macro-$F_1$ to account for class imbalance across solvent labels.
For host–guest description, we evaluate open-ended answers with Rouge-1 recall, precision, and $F_1$.
For the molecular identification task, we report SMILES validity, canonical-SMILES exact match, InChIKey first-block match, molecular-formula match, Morgan-fingerprint Tanimoto similarity, and heavy-atom count error ($\Delta$Heavy), so that exact recovery and chemically meaningful near misses are both visible.
Full definitions of these evaluation metrics are discussed in Appendix~\ref{app:metrics}.

\section{Experiments}
\label{sec:experiments}
We evaluate a broad range of LLMs, including open-sourced and proprietary model families, scales from 8B to frontier closed systems, and recent release dates. 
The open-weight set includes Qwen3.5-\{9B, 27B\}~\citep{qwen3.5} and Llama-3.1-\{8B, 70B\}-Instruct~\citep{grattafiori2024llama}, and DeepSeek-v4~\citep{deepseekai2026deepseekv4}. 
The proprietary set includes GPT-5.4-\{Mini, Nano\}~\citep{openai2026gpt5.4}, and Gemini-3-Flash~\citep{google2023gemini3}.
For a rigorous and fair evaluation and comparison, we leverage OpenRouter~\citep{openrouter} for all model inferences.  
Detailed discussion about experimental setup is provided in Appendix~\ref{app:implementation}.

\begin{table*}[t]
\centering
\small
\setlength{\tabcolsep}{6pt}
\renewcommand{\arraystretch}{0.9}
\resizebox{\linewidth}{!}{
\begin{tabular}{llccccccccc}
\toprule
{} & \multirow{2}{*}{\textbf{Model}}
  & \multicolumn{2}{c}{\textbf{Binding Affinity}}
  & \multicolumn{2}{c}{\textbf{Top-Binder}}
  & \multicolumn{2}{c}{\textbf{Solvent}}
  & \multicolumn{3}{c}{\textbf{Host–Guest Description}} \\
\cmidrule(lr){3-4}\cmidrule(lr){5-6}\cmidrule(lr){7-8}\cmidrule(lr){9-11}
& & MAE $\downarrow$ & RMSE $\downarrow$ & ACC $\uparrow$ & Regret $\downarrow$ & F1 $\uparrow$ & B. Acc $\uparrow$ & Recall $\uparrow$ & Precision $\uparrow$ & F1 $\uparrow$ \\
\midrule
\multirow{8}{*}{\rotatebox{90}{Base}}
   & Qwen3.5-9B     & $2.491$ & $3.360$ & $0.379$ & $0.930$ & $0.159$ & $0.166$ & $0.040$ & $0.023$ & $0.043$ \\
   & Qwen3.5-27B    & $1.803$ & $2.503$ & $0.404$ & $0.851$ & $0.225$ & $0.364$ & $0.495$ & $0.072$ & $0.122$ \\
   & Llama3.1-8B    & $2.699$ & $3.630$ & $0.228$ & $1.281$ & $0.151$ & $0.225$ & $0.266$ & $0.059$ & $0.092$ \\
   & Llama3.1-70B   & $1.632$ & $2.149$ & $0.338$ & $1.054$ & $0.118$ & $0.254$ & $0.487$ & $\mathbf{0.091}$ & $\mathbf{0.152}$ \\
   & GPT-5.4-Mini   & $1.549$ & $2.182$ & $0.428$ & $0.810$ & $0.219$ & $0.274$ & $0.437$ & $0.086$ & $0.137$ \\
   & GPT-5.4-Nano   & $1.642$ & $2.169$ & $0.411$ & $0.816$ & $0.182$ & $0.347$ & $0.472$ & $0.062$ & $0.107$ \\
   & Gemini-3-Flash & $\mathbf{1.248}$ & $\mathbf{1.679}$ & $\mathbf{0.498}$ & $\mathbf{0.647}$ & $\mathbf{0.350}$ & $\mathbf{0.470}$ & $\mathbf{0.506}$ & $0.067$ & $0.118$ \\
   & DeepSeek-v4    & $\underline{1.433}$ & $\underline{1.994}$ & $\underline{0.461}$ & $\underline{0.730}$ & $\underline{0.309}$ & $\underline{0.381}$ & $\underline{0.500}$ & $\underline{0.090}$ & $\underline{0.141}$ \\
\midrule
\multirow{8}{*}{\rotatebox{90}{Few-Shot}}
   & Qwen3.5-9B     & $3.650$ & $4.820$ & $0.370$ & $0.951$ & $0.154$ & $0.150$ & $0.000$ & $0.022$ & $0.042$ \\
   & Qwen3.5-27B    & $2.258$ & $3.256$ & $0.392$ & $0.889$ & $0.178$ & $0.257$ & $0.636$ & $\mathbf{0.585}$ & $\mathbf{0.580}$ \\
   & Llama3.1-8B    & $5.504$ & $6.940$ & $0.283$ & $1.227$ & $0.142$ & $0.182$ & $0.655$ & $0.369$ & $0.456$ \\
   & Llama3.1-70B   & $1.774$ & $2.359$ & $0.354$ & $1.026$ & $0.144$ & $0.185$ & $0.631$ & $\underline{0.474}$ & $\underline{0.531}$ \\
   & GPT-5.4-Mini   & $1.958$ & $2.808$ & $0.430$ & $0.824$ & $0.141$ & $\underline{0.291}$ & $0.542$ & $0.228$ & $0.307$ \\
   & GPT-5.4-Nano   & $2.176$ & $2.894$ & $0.419$ & $0.819$ & $0.190$ & $0.270$ & $0.532$ & $0.095$ & $0.152$ \\
   & Gemini-3-Flash & $\mathbf{1.257}$ & $\mathbf{1.702}$ & $\mathbf{0.513}$ & $\mathbf{0.619}$ & $\mathbf{0.389}$ & $\mathbf{0.421}$ & $\underline{0.660}$ & $0.364$ & $0.448$ \\
   & DeepSeek-v4    & $\underline{1.618}$ & $\underline{2.276}$ & $\underline{0.470}$ & $\underline{0.713}$ & $\underline{0.203}$ & $0.225$ & $\mathbf{0.720}$ & $0.303$ & $0.352$ \\
\midrule
\multirow{8}{*}{\rotatebox{90}{CoT}}
   & Qwen3.5-9B     & $3.664$ & $4.885$ & $0.382$ & $0.944$ & $0.167$ & $0.197$ & $0.300$ & $0.039$ & $0.068$ \\
   & Qwen3.5-27B    & $2.438$ & $3.468$ & $0.398$ & $0.898$ & $0.254$ & $\underline{0.415}$ & $\mathbf{0.526}$ & $0.051$ & $0.092$ \\
   & Llama3.1-8B    & $4.911$ & $6.279$ & $0.293$ & $1.220$ & $0.154$ & $0.153$ & $0.380$ & $\mathbf{0.102}$ & $\mathbf{0.144}$ \\
   & Llama3.1-70B   & $1.833$ & $2.512$ & $0.373$ & $0.985$ & $0.106$ & $0.380$ & $0.421$ & $0.055$ & $0.097$ \\
   & GPT-5.4-Mini   & $2.036$ & $2.887$ & $0.429$ & $0.828$ & $0.220$ & $0.282$ & $0.444$ & $0.080$ & $0.129$ \\
   & GPT-5.4-Nano   & $2.160$ & $2.881$ & $0.410$ & $0.822$ & $0.174$ & $0.257$ & $0.492$ & $0.056$ & $0.098$ \\
   & Gemini-3-Flash & $\mathbf{1.261}$ & $\mathbf{1.723}$ & $\mathbf{0.510}$ & $\mathbf{0.609}$ & $\mathbf{0.331}$ & $\mathbf{0.432}$ & $0.512$ & $0.062$ & $0.110$ \\
   & DeepSeek-v4    & $\underline{1.541}$ & $\underline{2.183}$ & $\underline{0.445}$ & $\underline{0.743}$ & $\underline{0.307}$ & $0.414$ & $\underline{0.522}$ & $\underline{0.080}$ & $\underline{0.134}$ \\
\bottomrule
\end{tabular}

}
\caption{Main performance comparison across the four fundamental tasks of \modelname.
For each setting, the best score is shown in $\mathbf{bold}$, and the second-best is $\underline{\text{underlined}}$.
}
\label{tab:main_results}
\end{table*}

\subsection{Main Results}
Table~\ref{tab:main_results} reports the performance on \modelname over eight LLMs under three prompting strategies.
According to the table, we conclude that:
(i) Prompting strategy effects are highly task-dependent.
Specifically, Few-Shot improves the performance of the host–guest description task for every model, while downgrading on binding affinity prediction, compared with Base.
Moreover, CoT consistently underperforms Few-Shot and Base on the host–guest description task.
This observation shows that \textit{no single prompting strategy uniformly improves performance across various tasks.}
We analyze this phenomenon in Section~\ref{sec:exp:case_study}.
(ii) Gemini-3-Flash achieves the best score on binding affinity prediction, top-binder selection, and solvent identification across every prompting setting, with DeepSeek-v4 consistently in second place and GPT-5.4-Mini close behind.
This observation shows that \textit{frontier proprietary LLMs deliver state-of-the-art on supramolecular chemistry tasks.}

\subsection{Molecular Identification}
\label{sec:exp:vision}

We additionally evaluate \emph{molecular identification}, and visualize the results in Figure~\ref{fig:molecular_identification}.
The full numeric values are reported in Appendix Table~\ref{tab:vision}.
According to Figure~\ref{fig:molecular_identification}, we observe that:
(i) Gemini-3-Flash dominates every column under Few-Shot, while GPT-5.4-Nano collapses to $\leq 0.11$ Canonical in every prompting setting despite emitting valid SMILES on a majority of prompts.
This observation shows that \textit{frontier multimodal training yields a large image-to-SMILES gap that smaller proprietary models fail to close}.
(ii) Across all models, the gap between Canonical and InChIKey or Tanimoto remains large, indicating that models recover the molecular scaffold roughly right (high Tanimoto) but miss the exact connectivity (low Canonical).
This observation demonstrates that \textit{the visual chemistry knowledge of current LLMs is partial rather than absent.}
(iii) CoT consistently degrades identification quality, \textit{e.g.}, Gemini-3-Flash's Canonical drops from $0.593$ under Few-Shot to $0.567$ under CoT, suggesting that \textit{an explicit reasoning cue destabilizes image-grounded molecular translation rather than improves it.}

\begin{figure*}[t]
    \centering
    \includegraphics[width=\linewidth]{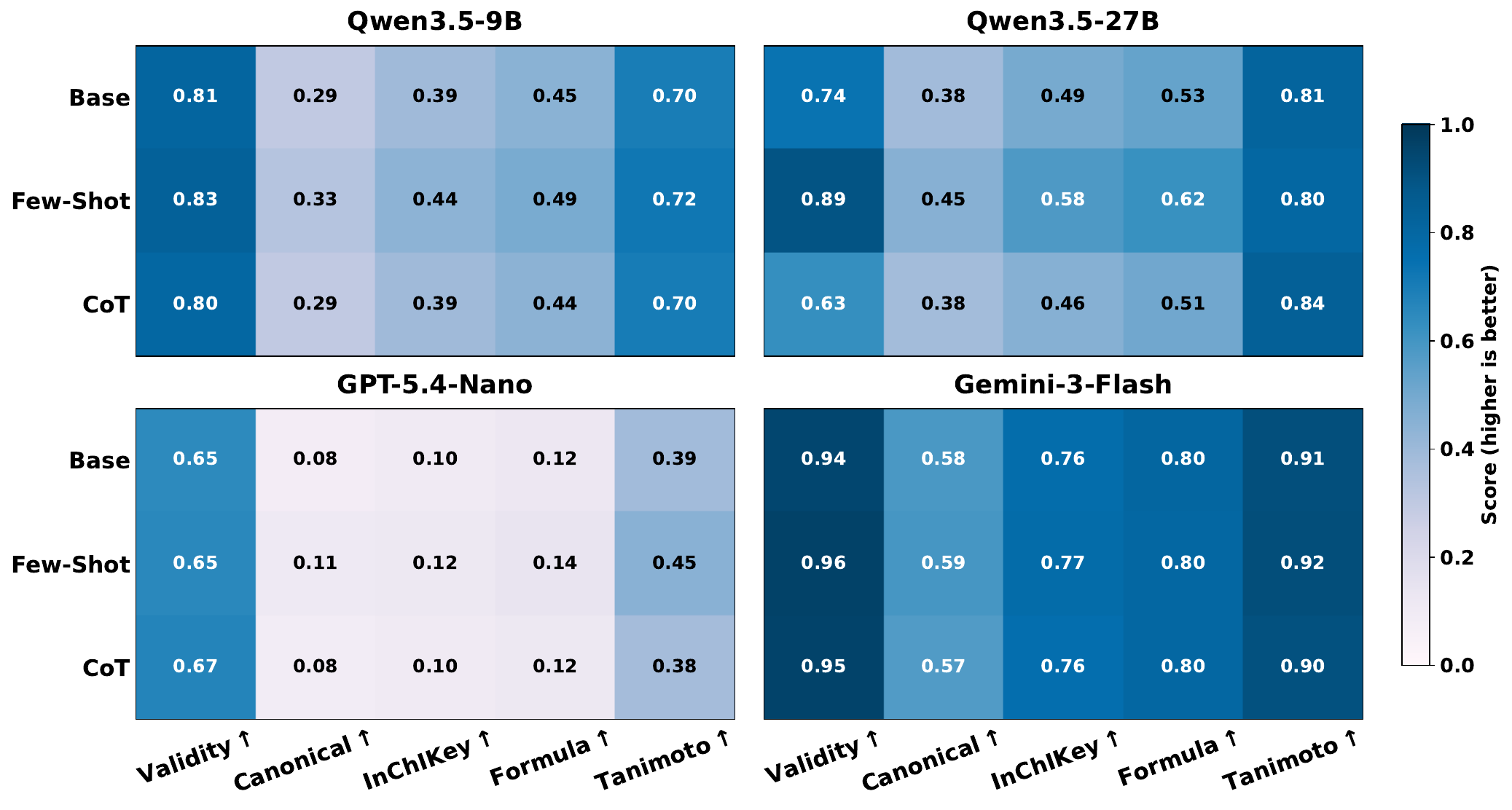}
    \caption{Heatmap for molecular identification task (full numeric values in Table~\ref{tab:vision}).}
    \label{fig:molecular_identification}
\end{figure*}

\subsection{Domain Adaptation Analysis}
\label{sec:domain_adapt}
We apply DAPT~\citep{ibrahim2024simple, gururangan2020don} to train two open-weight small models, \textit{i.e.}, Qwen3.5-9B and Llama3.1-8B, on our constructed supramolecular chemistry text corpus \corpusname and evaluate them in all four fundamental tasks under the Base prompting setting.
The details of the implementation are provided in Appendix~\ref{app:dapt_recipe}, and Table~\ref{tab:domain_adaptation} lists the result.
According to Table~\ref{tab:domain_adaptation}, we observe that:
(i) DAPT substantially improves binding affinity prediction performance for both models, which validates that the \textit{\corpusname directly transfers to the in-distribution binding affinity regression task.}
(ii) DAPT also lifts the model performance in host–guest description tasks, with the smaller Llama3.1-8B showing larger absolute gains than Qwen3.5-9B on most columns.
This validates that \textit{\corpusname exposure transfers beyond regression into open-ended generation}.
(iii) The performance on the top-binder selection task downgrades for both models, and the Llama variant additionally fails to follow the strict letter format on solvent identification, showing that \textit{DAPT on free-form scientific text trades off against strict letter-format MCQ output.}

\begin{table}[t]
\centering
\small
\setlength{\tabcolsep}{1pt}
\resizebox{\linewidth}{!}{
\begin{tabular}{lccccccc}
\toprule
\multirow{2}{*}{\textbf{Setting}}
  & \multicolumn{2}{c}{\textbf{BAP}}
  & \textbf{TBS}
  & \textbf{SID}
  & \multicolumn{3}{c}{\textbf{HGD}} \\
\cmidrule(lr){2-3}\cmidrule(lr){4-4}\cmidrule(lr){5-5}\cmidrule(lr){6-8}
& MAE $\downarrow$ & RMSE $\downarrow$ & ACC $\uparrow$ & F1 $\uparrow$ & Rec. $\uparrow$ & Prec. $\uparrow$ & F1 $\uparrow$ \\
\midrule
Qwen3.5-9B             & $2.491$          & $3.360$          & $\mathbf{0.379}$ & $0.159$          & $0.040$ & $0.023$ & $0.043$ \\
$+$ \corpusname        & $\mathbf{2.173}$ & $\mathbf{2.737}$ & $0.235$          & $\mathbf{0.161}$ & $\mathbf{0.053}$      & $\mathbf{0.048}$      & $\mathbf{0.050}$      \\
Llama3.1-8B            & $2.699$          & $3.630$          & $\mathbf{0.228}$ & $\mathbf{0.151}$ & $0.266$ & $0.059$ & $0.092$ \\
$+$ \corpusname        & $\mathbf{1.636}$ & $\mathbf{2.204}$ & $0.152$          & --         & $\mathbf{0.311}$      & $\mathbf{0.074}$      & $\mathbf{0.106}$      \\
\bottomrule
\end{tabular}
}
\caption{Performance of DAPT on \corpusname for Qwen3.5-9B and Llama3.1-8B.}
\label{tab:domain_adaptation}
\end{table}

\subsection{Per-Host Analysis}
\label{sec:exp:host_breakdown}

\begin{figure*}[t]
    \centering
    \includegraphics[width=0.98\linewidth]{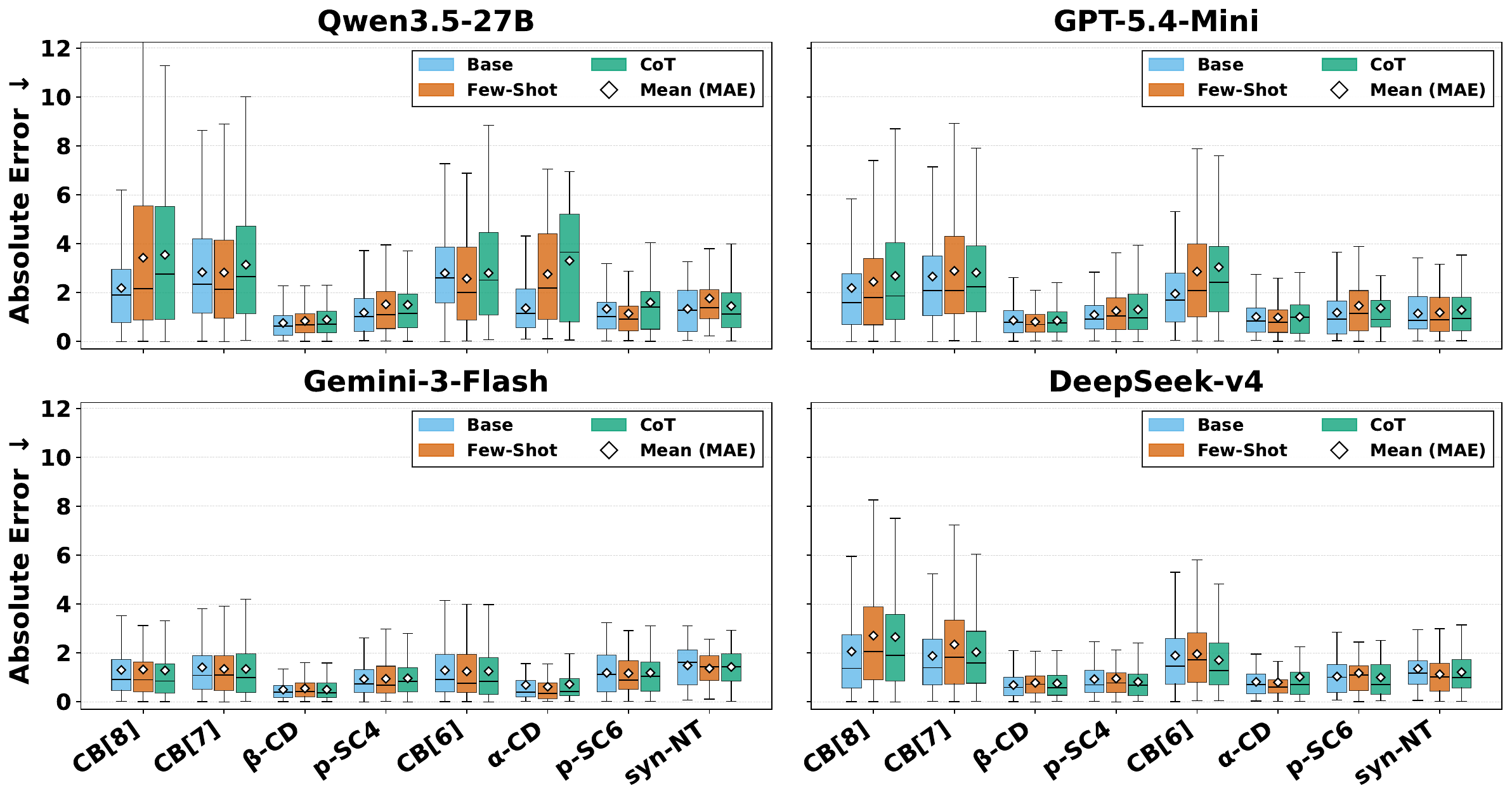}
    \caption{Absolute error distributions for binding affinity prediction grouped by the eight most frequent hosts.
    Note that CB[$n$] represents Cucurbit[$n$]uril, $\alpha/\beta$-CD denotes $\alpha/\beta$-Cyclodextrin, $p$-SC$n$ is $p$-Sulfonatocalix[$n$]arene, and syn-NT means syn-Amide Naphthotube.}
    \label{fig:per_host_mae_task1}
\end{figure*}

To understand the performance breakdown in the binding affinity prediction task, we conduct an experiment to evaluate the results grouped by the eight most frequent hosts, and report the results in Figure~\ref{fig:per_host_mae_task1}.
According to Figure~\ref{fig:per_host_mae_task1}, we observe that:
(i) $\beta$-CD is the easiest host for every setting, while $18$ of the $20$ worst cells fall on CB[$n$], showing that \textit{per-host difficulty is highly heterogeneous and CB[$n$] hosts drive the headline error.}
(ii) Switching from Base to Few-Shot or CoT leaves the performance of Gemini-3-Flash's error remain unchanged, and degrades the other three models, \textit{i.e.}, Qwen3.5-27B, GPT-5.4-Mini, and DeepSeek-v4.
This observation shows that the prompting strategy interacts strongly with base-model capacity rather than uniformly improving it.

\subsection{Why Prompt Strategies Hurt?}
\label{sec:exp:case_study}

To diagnose the CoT regressions reported in Table~\ref{tab:main_results}, we walk through a representative failure of DeepSeek-v4 on binding affinity prediction, with the full prompt and traces shown in Figure~\ref{fig:case_study_cot}.
For the host–guest pair (\emph{4,7,13,18-Tetraoxa-1,10-diazabicyclo[8.5.5]icosane}, \emph{Ba$^{2+}$}) with reference $\log K_a = 2.00$, the Base prompt yields a near-perfect prediction of $2.10$, while CoT yields $11.00$, off by nine orders of magnitude in the association constant.
Inspecting the CoT trace reveals the failure mode: the model confidently asserts that ``the binding constant for Ba$^{2+}$ with such bicyclic diaza-crown ethers typically falls in the range of $\log K_a \approx 10$--$12$'' and lands on $11.0$ as a ``reasonable estimate'' from ``typical literature values for similar cryptands'', while the actual literature value sits near $2.0$.

This failure mode generalizes the prompt-strategy regression observed across binding affinity prediction and molecular identification.
CoT prompts the model to articulate a reasoning chain, but articulation is not knowledge, \textit{i.e.}, when invited to cite ``typical literature values'', the model fabricates a confident range that does not exist, and the qualitative chemistry it does recover (preorganized cavity, favorable ion–dipole interactions) has no calibrated mapping to $\log K_a$ magnitudes.
In short, \textit{CoT only helps when the model can actually reason about the underlying chemistry, and supramolecular reasoning at scientist-level rigor remains beyond the reach of current LLMs.}

\section{Insights}
\label{sec:insights}
We close by distilling the key insights from \modelname throughout the paper.

\paragraph{\textcircled{\small 1}~Frontier scale dominates, but every task leaves headroom.}
Gemini-3-Flash leads binding affinity prediction, top-binder selection, and solvent identification under every prompting strategy (Table~\ref{tab:main_results}), confirming that frontier proprietary multimodal training transfers to supramolecular reasoning.
Even so, the best $\log K_a$ MAE remains $1.25$, the best top-binder accuracy plateaus at $0.513$, and host–guest description Rouge-1 $F_1$ stays below $0.6$ for every model, leaving substantial room for improvement on every task.

\paragraph{\textcircled{\small 2}~No prompting strategy is universally helpful.}
Few-Shot improves host–guest description for nearly every model but degrades binding affinity prediction, \textit{e.g.}, Llama3.1-8B inflates from MAE $2.699$ to $5.504$, and CoT consistently hurts binding affinity prediction and molecular identification.
The effect is mediated by base-model capacity, \textit{i.e.}, Gemini-3-Flash is largely robust across prompting strategies, while smaller models pay a noticeable penalty.
Practitioners should therefore choose prompting strategies per task and per base model, not as a one-size-fits-all default.

\paragraph{\textcircled{\small 3}~CoT amplifies, rather than fixes, the underlying reasoning gap.}
The case study in Section~\ref{sec:exp:case_study} shows that when a model lacks the supramolecular knowledge to ground its qualitative chemistry in $\log K_a$ magnitudes, asking it to articulate a reasoning chain yields fluent but uncalibrated text and chemically nonsensical predictions.
CoT is therefore not a substitute for domain knowledge in supramolecular chemistry, and targeted domain adaptation or external chemistry tools are likely more productive directions for closing the gap.

\paragraph{\textcircled{\small 4}~Domain adaptation has uneven transfer.}
A single DAPT recipe on the \corpusname corpus improves binding affinity prediction MAE for both adapted models, \textit{i.e.}, Qwen3.5-9B and Llama3.1-8B (Table~\ref{tab:domain_adaptation}), but degrades top-binder selection accuracy for both, and the Llama variant collapses on solvent identification due to format-following failures.
Free-form scientific text adaptation transfers cleanly to regression but not to strict letter-format MCQ output, suggesting that domain-adaptive pretraining and instruction-format preservation need to be optimized jointly.

\paragraph{\textcircled{\small 5}~Molecular identification recovers scaffolds but not exact connectivity.}
On the molecular identification task (Figure~\ref{fig:molecular_identification}), every model shows a large gap between Canonical SMILES match and the looser InChIKey or Tanimoto similarity metrics, indicating that models recover the molecular scaffold roughly right but miss the exact connectivity.
This split signals that the bottleneck is not abstract visual chemistry comprehension but precise bond-level reasoning, an axis on which CoT appears to degrade performance rather than improve it.

\section{Conclusion}
We introduce the first host–guest supramolecular chemistry benchmark for LLMs, called \modelname, that contains four fundamental tasks, \textit{i.e.}, binding affinity prediction, top-binder selection, solvent identification, and host–guest description, along with an auxiliary vision-based task for molecular identification.
Moreover, we release a 16M-token text corpus, \corpusname, comprising Europe PMC supramolecular chemistry articles that can contribute future studies in this domain.
We evaluate eight LLMs and reveal several key findings:
(i) frontier proprietary LLMs deliver SOTA, yet every task leaves substantial headroom for improvement; (ii) no single prompting strategy is universally beneficial, with CoT in particular amplifying rather than fixing the underlying reasoning gap when the model lacks supramolecular knowledge; (iii) DAPT over our released \corpusname text corpus transfers strongly to in-distribution regression but trades off against strict letter-format MCQ output; and (iv) multimodal models recover molecular scaffolds from 2D structural drawings but miss exact bond-level connectivity, with CoT degrading rather than improving identification quality.
These results highlight that supramolecular chemistry remains a genuine bottleneck for current LLMs, and we hope \modelname catalyzes future research on chemistry-grounded LLMs.
\section*{Limitations}

\paragraph{Public-data memorization risk.}
In this study, we leverage SupraBank and Europe PMC for benchmarking and text corpus construction, which are both publicly available resources.
Frontier proprietary LLMs may have seen part of this distribution during pretraining, so the absolute performance numbers may overestimate true generalization to genuinely novel host–guest pairs.
We do not run a controlled novel host or temporal-cut split to quantify this effect.

\paragraph{Closed-API drift.}
Proprietary models are accessed \textit{via} OpenRouter, which does not pin model versions across requests.
As providers update underlying checkpoints, the absolute numbers we report may drift over time.

\section*{Ethical Consideration}

\paragraph{Data and licensing.}
The host–guest binding records are derived from SupraBank under its public-distribution terms, and the \corpusname text corpus is built only from open-access Europe PMC articles, subject to each article's individual license.
The benchmark contains no human-subjects data and no personally identifiable information.

\paragraph{Intended use.}
\modelname is a research benchmark for evaluating LLM reasoning on host–guest supramolecular chemistry.
It is not intended to drive clinical, regulatory, or production drug-design decisions without expert oversight and experimental validation.

\paragraph{Fabrication risk.}
LLMs can produce confident, fluent, but chemically incorrect predictions under CoT prompting.
Downstream users should treat model outputs as hypotheses to be verified against the literature and against established computational or experimental methods, especially when binding affinity magnitudes inform candidate selection.

\paragraph{Reproducibility.}
We release task data, evaluation code, prompts, and the DAPT recipe.
For the proprietary models accessed \textit{via} OpenRouter, exact reproduction depends on provider-side version stability beyond our control, and we record the model identifier and request date with each inference run.

\bibliography{reference}

\appendix
\titlecontents{section}[0pt]
  {\addvspace{1.4ex}\large\bfseries}
  {\thecontentslabel\hspace{1em}}
  {\hspace*{-1em}}
  {\titlerule*[0.5em]{.}\contentspage}
\titlecontents{subsection}[2.4em]
  {\addvspace{0.5ex}\normalsize}
  {\thecontentslabel\hspace{0.8em}}
  {\hspace*{-2em}}
  {\titlerule*[0.5em]{.}\contentspage}

\startcontents[appendix]
\onecolumn
\begin{center}
  {\LARGE\bfseries Appendix Contents}
\end{center}
\vspace{1.5em}
\printcontents[appendix]{}{1}{}
\vspace{1.5em}
\twocolumn
\section{Detailed Related Works}
\label{app:related_works}

\subsection{Foundation Models for Chemistry}

Although AI-driven methods \cite{zhao2023self,ju2022grape,qian2022co,zhao2021multi,ju2023graphpatcher} have long been applied to a wide range of chemistry problems, recent advances in LLMs have significantly expanded the scope of what AI systems can achieve in chemical understanding and reasoning. Foundation models powered by LLMs can now read, write, and reason about chemistry with growing fluency~\citep{wang2025generative,wang2025beyond,wang2024gft,zhao2026controllable,wang2026molecular}.
Early efforts treated chemistry as one slice of a broader scientific corpus: Galactica pretrained a single model on millions of papers, textbooks, and reference works, and showed that a generalist LM can recall chemical facts and manipulate SMILES strings with surprising fidelity \citep{taylor2022galactica}.
A parallel line of work has framed chemistry as translation between natural language and molecular notation: MolT5 \citep{edwards2022molt5} jointly pre-trains a sequence-to-sequence model on text and SMILES, and nach0 \citep{livne2024nach0} extends this idea into a unified multitask foundation model that handles named entity recognition, property prediction, and forward and retro reaction prediction within a single decoder.
Building on these foundations, subsequent work has pushed in two complementary directions.
The first is {domain-adapted LLMs} that continue training on chemistry-specific corpora and instruction data; representative examples include ChemLLM, ChemDFM, LlaSMol, and the Mol-Instructions resource, which together show that targeted adaptation of mid-scale open models (7--13B) can match or exceed much larger general-purpose models on molecular description, property prediction, and reaction tasks \citep{zhang2024chemllm,zhao2025chemdfm,yu2024llasmol,fang2024molinstructions}.
The second is {LLM-driven chemistry agents} that couple an LLM controller to external tools and laboratory automation.
ChemCrow \citep{bran2024chemcrow} interfaces a planner with retrosynthesis, property-prediction, and web-search modules to execute multi-step chemical workflows; Coscientist \citep{boiko2023coscientist} drives a cloud lab end-to-end, autonomously planning and running palladium-catalysed cross-coupling optimisations; and ChemAgent \citep{tang2025chemagent} maintains a self-updating set of planning, execution, and knowledge memories that improves chemical reasoning as the agent accumulates experience.
Despite this progress, existing systems are almost exclusively evaluated on {small-molecule} tasks such as single-molecule property prediction, retrosynthesis, and reaction-yield estimation.
None target {supramolecular} reasoning, which involves a pair (or set) of molecules interacting through non-covalent forces and therefore stresses different capabilities, including host–guest geometry, binding thermodynamics, and application-aware design.
\modelname{} addresses this gap with tasks that span affinity prediction, host–guest description, application-level inference, and property-conditioned generation.

\subsection{Chemistry Benchmarks}
Benchmarks have been a central driver of progress in machine-learning chemistry.
For graph- and fingerprint-based deep models, {MoleculeNet} \citep{wu2018moleculenet} unified property-prediction tasks across quantum mechanics, physiology, and toxicity, {Therapeutics Data Commons} \citep{huang2021tdc} extended the same idea to drug-discovery pipelines, and {GuacaMol} \citep{brown2019guacamol} and {MOSES} \citep{polykovskiy2020molecular} standardized the evaluation of generative models for de novo molecule design.
For LLMs specifically, {ChemLLMBench} \citep{guo2023chemllmbench} provided an early eight-task suite spanning name prediction, property classification, and reaction prediction, and {ChemBench} \citep{mirza2025chembench} subsequently scaled the idea to over 2{,}700 expert-annotated questions and showed that frontier proprietary models can rival or exceed expert chemists on textbook-style queries.
Broader scientific suites complement these chemistry-specific resources: {SciBench} and {SciEval} \citep{wang2024scibench,sun2024scieval} probe quantitative reasoning across chemistry, physics, and biology, while {LAB-Bench} \citep{laurent2024labbench} evaluates research-assistant capabilities such as literature search, protocol planning, and figure interpretation in the life sciences.
Across this landscape, however, every benchmark targeted at LLMs focuses on {single-molecule} or {single-system} reasoning: property prediction for one molecule, generation of one molecule, or knowledge questions about a single concept or protocol.
The closest analog on the molecular-modeling side is the SAMPL host–guest blind challenge \citep{amezcua2022sampl8}, but it is designed to calibrate physics-based free-energy methods on a handful of curated pairs rather than to evaluate language models, and it does not cover application-level outcomes such as drug-delivery vehicle selection, sensor design, or toxin sequestration.
\modelname{} fills this gap by evaluating LLMs on supramolecular host–guest reasoning under a unified protocol, with task families that span both foundational competencies and application-level inference.

\subsection{Supramolecular Chemistry}
\label{subsec:supra}
Supramolecular chemistry studies molecular assemblies held together not by covalent bonds but by reversible non-covalent interactions, including hydrogen bonding, hydrophobic effects, $\pi$-stacking, electrostatic forces, and metal coordination \citep{lehn1995supramolecular,steed2009supramolecular}.
A typical system pairs a {host}, often a macrocycle with a well-defined cavity, with a {guest} bound through molecular recognition \citep{ariga2012molecular}.
The dominant host families in current practice include cyclodextrins, cucurbiturils, calixarenes, pillararenes, and crown ethers, as well as extended porous architectures such as metal-organic and covalent-organic frameworks (MOFs and COFs).
These platforms underpin three major clinical applications of supramolecular chemistry.
In {drug delivery}, cyclodextrin-based formulations improve the solubility and bioavailability of hydrophobic active ingredients \citep{loftsson2010cyclodextrins}.
In {sensing}, host–guest recognition coupled with optical or electrochemical reporters enables detection of analytes ranging from pollutants to disease biomarkers \citep{you2015sensing}.
In {toxin sequestration}, macrocyclic receptors are being developed as in-vivo antidotes for drugs of abuse and other toxins, building on the sugammadex precedent \citep{deng2020sequestration}; an acyclic pillararene host, for example, sequesters both methamphetamine and fentanyl {in vivo} and reverses their pharmacological effects \citep{brockett2023pillar6maxq}.

The same properties that make supramolecular systems powerful also make them difficult to model.
Binding affinities are highly sensitive to solvent, pH, ionic strength, and counter-ion identity, so values reported by different laboratories are often not directly comparable.
Both hosts and guests can adopt multiple conformations, and a single pair may bind in more than one geometry (\textit{e.g.}, inclusion vs.\ external complexation), which complicates experimental fitting and computational simulation alike.
High-fidelity tools such as DFT and molecular dynamics can in principle resolve these ambiguities, but they are slow, expensive, and require expert setup; even in the carefully curated SAMPL host–guest blind challenges, leading free-energy methods still incur RMSE on the order of 1--2~kcal/mol on small benchmark sets \citep{amezcua2022sampl8}.
As a consequence, curated binding-affinity datasets remain small and fragmented relative to the small-molecule property-prediction datasets that drive mainstream ML chemistry, and recent reviews have identified data scarcity and heterogeneity as the central bottleneck for machine learning on supramolecular systems \citep{colaco2024supraml}.
\modelname is designed to expose and quantify these difficulties for modern LLMs.

\section{Implementation Details}
\label{app:implementation}

\subsection{Environment}
All experiments are conducted on a Linux OS equipped with four Nvidia A100 GPUs. The models are implemented using PyTorch 2.4.0 with CUDA 12.1 and Python 3.11.5.

\subsection{Evaluation Metrics}
\label{app:metrics}
In this section, we discuss the employed evaluation metrics in \modelname.
For a test sample $i$, $\hat{y}_i$ and $y_i$ denote the extracted model prediction and the gold reference, respectively. 

\paragraph{MAE and RMSE.}
For scalar predictions $\hat{y}_i$ against scalar gold references $y_i$,
\begin{align}
    \mathrm{MAE} \;&=\; \frac{1}{N}\sum_{i=1}^{N} |\hat{y}_i - y_i|, \\
    \mathrm{RMSE} \;&=\; \sqrt{\frac{1}{N}\sum_{i=1}^{N} (\hat{y}_i - y_i)^2}.
\end{align}

\paragraph{Letter Accuracy.}
For categorical predictions $\hat{\ell}_i$ drawn from a discrete option set $\{A, B, C, D, \ldots\}$ against gold labels $\ell_i$,
\begin{equation}
\mathrm{ACC} \;=\; \frac{1}{N}\sum_{i=1}^{N} \mathbb{1}\!\left[\hat{\ell}_i = \ell_i\right].
\end{equation}

\paragraph{Binding-Affinity Regret.}
For the top-binder selection task, accuracy treats every wrong option equally even when some wrong choices are close to the strongest binder.
Let $\mathcal{O}_i$ denote the candidate option set for item $i$, let $k_i^\star=\arg\max_{k\in\mathcal{O}_i}\log K_{a,i}^{(k)}$ be the strongest candidate, and let $\hat{k}_i$ be the model-selected candidate.
We define binding-affinity regret as the mean loss in $\log K_a$ caused by selecting $\hat{k}_i$ instead of $k_i^\star$,
\begin{equation}
\mathrm{Regret} \;=\; \frac{1}{N}\sum_{i=1}^{N}
\left(\log K_{a,i}^{(k_i^\star)} - \log K_{a,i}^{(\hat{k}_i)}\right).
\end{equation}
This metric distinguishes chemically near-optimal mistakes from selections that miss the strongest binder by several orders of magnitude.

\paragraph{Macro-$F_1$.}
For multi-class classification over a class set $\mathcal{C}$, Macro-$F_1$ is the unweighted mean of per-class $F_1$,
\begin{equation}
\mathrm{Macro\text{-}}F_1 \;=\; \frac{1}{|\mathcal{C}|}\sum_{c\in\mathcal{C}} \frac{2\,P_c\,R_c}{P_c + R_c},
\end{equation}
where $P_c$ and $R_c$ are the per-class precision and recall.

\paragraph{Balanced Accuracy.}
For the solvent identification task, the label distribution is imbalanced because aqueous measurements dominate the dataset.
We therefore also report balanced accuracy, defined as the unweighted mean of per-class recall over the solvent class set $\mathcal{C}$,
\begin{equation}
\mathrm{Balanced\text{-}ACC} \;=\; \frac{1}{|\mathcal{C}|}\sum_{c\in\mathcal{C}} R_c.
\end{equation}
Balanced accuracy penalizes majority-class collapse and directly measures whether a model recovers minority solvent regimes, rather than only exploiting the dominant water prior.

\paragraph{Rouge-1 Recall, Precision, and $F_1$.}
Let $T(\cdot)$ denote the unigram token bag of a text after Porter stemming and case folding. For a prediction $\hat{a}_i$ and gold answer $a_i$, the per-row Rouge-1 quantities are
\begin{align}
R_i &= \frac{|T(\hat{a}_i)\cap T(a_i)|}{|T(a_i)|},\\
P_i &= \frac{|T(\hat{a}_i)\cap T(a_i)|}{|T(\hat{a}_i)|},\\
F_{1,i} &= \frac{2\,P_i\,R_i}{P_i + R_i}.
\end{align}
We report the unweighted means $\overline{R}$, $\overline{P}$, $\overline{F_1}$ across the $N$ examples, and empty predictions contribute zero to all three.

\paragraph{Validity.}
This metric computes the fraction of predicted SMILES strings $\hat{s}_i$ that parse successfully~\citep{polykovskiy2020molecular, brown2019guacamol, weininger1988smiles}. 
Let $\mathrm{Mol}(\cdot)$ denote RDKit's SMILES parser, where $\mathrm{Mol}(s)=\bot$ indicates a parse failure, validity is computed as:
\begin{equation}
\mathrm{Validity} = \frac{1}{N}\sum_{i=1}^{N} \mathbb{1}[\mathrm{Mol}(\hat{s}_i)\neq\bot].
\end{equation}
We additionally define $\mathcal{V}=\{i:\mathrm{Mol}(\hat{s}_i)\neq\bot \wedge \mathrm{Mol}(s_i)\neq\bot\}$ as the set of examples on which both the prediction and the reference parse, used by Tanimoto and $\Delta$Heavy below.

\paragraph{Canonical-SMILES Exact Match.}
Canonical~\citep{schneider2015get, weininger1989smiles} is the strictest correctness criterion, \textit{i.e.}, a prediction counts only when its RDKit canonical SMILES coincides with the gold,
\begin{equation}
\mathrm{Canonical} = \frac{1}{N}\sum_{i=1}^{N} \mathbb{1}\!\left[\mathrm{Canon}(\hat{s}_i) = \mathrm{Canon}(s_i)\right].
\end{equation}
This metric rewards recovery of the exact constitution, bond orders, and (where written) stereochemistry of the reference; predictions that differ only in non-semantic SMILES variation (atom ordering, branch direction, aromaticity notation) canonicalize to the same form and are scored correctly.

\paragraph{InChIKey First-Block Match.}
InChIKey relaxes Canonical along the stereo and isotope dimensions~\citep{heller2015inchi, heller2013inchi}. 
Let $\mathrm{IK}(s)$ denote the first $14$-character block of the InChIKey of $s$, which is a hash over the molecule's main InChI layers (formula, connectivity, and fixed-hydrogen), so molecules that differ only in stereodescriptors or isotopic labeling collapse to the same prefix,
\begin{equation}
\mathrm{InChIKey} = \frac{1}{N}\sum_{i=1}^{N} \mathbb{1}[\mathrm{IK}(\hat{s}_i) = \mathrm{IK}(s_i)].
\end{equation}
The gap between Canonical and InChIKey quantifies how often the model recovers connectivity but misses stereo.

\paragraph{Molecular-Formula Match.}
Formula is the most lenient identity criterion, \textit{i.e.},  a prediction counts when its Hill-system molecular formula $F(\cdot)$ equals the gold.
We follow ~\citet{hill1900system} to implement this metric, which is computed as:
\begin{equation}
\mathrm{Formula} = \frac{1}{N}\sum_{i=1}^{N} \mathbb{1}[F(\hat{s}_i) = F(s_i)].
\end{equation}
A high Formula with low InChIKey indicates that the model reads atom counts off the image but cannot recover bonding.

\paragraph{Tanimoto Similarity.}
Tanimoto~\citep{morgan1965generation, bajusz2015tanimoto, rogers2010extended} is a soft structural similarity that smooths the binary identity metrics into a continuous score. 
For each $i\in\mathcal{V}$, let $f_i=\mathrm{FP}(\hat{s}_i)$ and $g_i=\mathrm{FP}(s_i)$ denote Morgan circular fingerprints with radius $r{=}2$ and $2048$-bit folding for the prediction and reference. 
Pairs where either fingerprint is undefined ($i\notin\mathcal{V}$) contribute $0$, so the metric is normalized over the full split,
\begin{equation}
\mathrm{Tanimoto} = \frac{1}{N}\sum_{i\in\mathcal{V}} \frac{|f_i\cap g_i|}{|f_i\cup g_i|}.
\end{equation}
This graded similarity lets a near-miss prediction (\textit{e.g.}, correct scaffold with one substituent wrong) earn partial credit it would not receive from exact-match metrics.

\paragraph{Heavy-Atom Count Difference ($\Delta$Heavy).}
$\Delta$Heavy~\citep{Landrum2016RDKit2016} measures how far the prediction deviates from the reference in molecular size, irrespective of connectivity. Let $H(\cdot)$ denote the number of non-hydrogen atoms. We report the mean absolute difference over parseable pairs,
\begin{equation}
\Delta\mathrm{Heavy} \;=\; \frac{1}{|\mathcal{V}|}\sum_{i\in\mathcal{V}} |H(\hat{s}_i) - H(s_i)|,
\end{equation}
in units of heavy atoms, with lower values being better.

\section{Dataset Construction}
\label{app:dataset}

We construct \modelname from experimentally reported host–guest binding records collected from SupraBank~\citep{suprabank}, a public repository\footnote{https://suprabank.org/} of peer-reviewed interactions in supramolecular chemistry.
We crawl each interaction together with its associated molecular metadata, \textit{i.e.}, host and guest names, identifiers, two-dimensional structure images, canonical SMILES strings, binding constants, solvent conditions, temperatures, pH, and the original literature citation, and de-duplicate by the SupraBank interaction ID.
The raw data contains $5{,}362$ measurements spanning $2{,}466$ unique molecular components from diverse supramolecular host families, including cucurbiturils, cyclodextrins, calixarenes, pillararenes, cavitands, and naphthotubes.
To obtain a clean, maintainable, and reproducible benchmark, we apply a six-step cleaning pipeline, and the final cleaned data contains $4{,}635$ records over $2{,}008$ components.

\paragraph{Step 1: Numeric Parsing.}
SupraBank stores binding constants, temperature, and pH as free-form strings with heterogeneous notation, \textit{e.g.}, \texttt{"7.76$\cdot$10$^{4}$"}, \texttt{"1.12$\cdot$10$^{7}$\, M$^{-1}$"}, \texttt{"25.0$^\circ$C"}. 
We implement a regex-based parser that normalizes these into numeric fields, \textit{i.e.}, $K_a$, $\log K_a$, and pH, handling scientific notation, units, and unicode multiplication marks.

\paragraph{Step 2: Organic-Solvent Filtering.}
Binding constants measured in organic solvents are thermodynamically incomparable to those measured in aqueous media. 
We discard records in which the field contains any of the common organic-solvent tokens, \textit{e.g.}, \{methanol, acetonitrile, DMSO, chloroform, dichloromethane, toluene, acetone, $\ldots$\}, and retain water-based systems, \textit{e.g.}, water, buffer, and D$_2$O. 

\paragraph{Step 3: Default-Condition Imputation.}
Many literature reports omit temperature and pH when measurements are taken under standard conditions. We impute missing values with the most-common literature defaults: $T = 25$\,$^\circ$C for $2{,}238$ records missing temperature, and $\mathrm{pH} = 7.0$ for $3{,}358$ records missing pH. 

\paragraph{Step 4: van't Hoff Temperature Correction.}
To make binding constants comparable across studies, we correct all $K_a$ values to a single reference temperature, \textit{i.e.}, $T_\mathrm{ref}=298.15$\, K, through the van't Hoff relation~\citep{atkins2023atkins, van1884etudes},
\begin{equation}
\ln\!\frac{K_a(T_\mathrm{ref})}{K_a(T)} \;=\; -\frac{\Delta H^\circ}{R}\!\left(\frac{1}{T_\mathrm{ref}} - \frac{1}{T}\right).
\end{equation}
Since most of the records are missing the $\Delta H^\circ$ field, we employ literature-averaged $\Delta H^\circ$ values for the most common hosts, \textit{e.g.}, CB[7] as $-40$\,kJ/mol, CB[8] as $-35$\,kJ/mol, $\beta$-CD as $-20$\,kJ/mol, \textit{etc.} 

\paragraph{Step 5: Per-Pair Averaging.}
For a measurement tuple, \textit{i.e.}, (\texttt{host}, \texttt{guest}, \texttt{pH}, \texttt{solvent}), we compute the geometric mean of $K_a$, which is equivalent to the arithmetic mean of $\log K_a$.
Note that we bin pH to $0.5$ units, \textit{e.g.}, pH $6.8$ and pH $7.1$ collapse to \texttt{pH bin}$=7.0$, while pH $6.0$ and pH $7.0$ remain distinct.

\paragraph{Step 6: Outlier Removal.}
Within each host–guest pair, we further employ Tukey's $1.5\,\mathrm{IQR}$ rule~\citep{tukey1977exploratory, hoaglin1987fine} on $\log K_a$ to flag outliers.
Specifically, a record is removed if it falls outside $[Q_1 - 1.5\,\mathrm{IQR},\, Q_3 + 1.5\,\mathrm{IQR}]$. 

\paragraph{Final Output.}
The cleaning pipeline above yields $4{,}635$ records over $2{,}008$ components
Each record carries $K_a$, $\log K_a$, and $\Delta G$ values at $25$\,$^\circ$C, full provenance fields (host/guest names, PubChem CIDs, SMILES, original literature citation), and a complete pipeline trail for reproducibility.

\section{Task Construction}
\label{app:task}

\subsection{Binding Affinity Prediction}
\label{app:bap}
Each row in the cleaned binding data, \textit{i.e.}, a host–guest pair with a numeric $\log K_a$ value and both host and guest SMILES available, becomes one regression question.
The prompt provides the host and guest names together with their canonical SMILES strings and asks the model to return a single $\log K_a$ value under standard aqueous conditions.
We hold out three exemplars drawn at the lowest, the median, and the highest $\log K_a$ in the pool as shared few-shot demonstrations and exclude them from the evaluation set.
The gold answer is preserved at four decimals, while few-shot demonstrations are rounded to one decimal so that the model is not biased toward a specific precision.
Random number generators are seeded for reproducibility.

\subsection{Top-Binder Selection}
\label{app:tbs}
We first drop hosts that admit fewer than four distinct guests.
Each question is then constructed by sampling four distinct guests uniformly at random from the remaining pool.
The sample is rejected if the spread between the highest and lowest $\log K_a$ falls below $0.5$, so that the gold answer leads the runner-up by a non-trivial margin, and rejected if the same four-guest set has already been sampled for this host.
The four candidates are randomly permuted across the answer letters A through D, and the correct letter is taken as the position of the maximum-$\log K_a$ guest among the displayed options.
We cap the number of questions per host at $200$ to prevent over-represented hosts from dominating the evaluation.
Three exemplars produced by the same procedure on the first three eligible hosts in alphabetical order are held out as shared few-shot demonstrations and excluded from the evaluation set.

\subsection{Solvent Identification}
\label{app:sid}
Each raw solvent string is mapped to one of six canonical classes (water, DMSO, MeCN, MeOH, CHCl$_3$, CH$_2$Cl$_2$) via a curated synonym map that absorbs deuterated and salt-buffered variants such as D$_2$O, buffer, DMSO-d$_6$, and CD$_2$Cl$_2$, and rows whose primary solvent reads ``complex'' are rescued through the secondary listing where possible; rows that match no canonical class are dropped.
Each question is a six-way multiple-choice question with the candidate solvent classes in fixed letter order A through F, and the prompt embeds a short domain guidance text relating qualitative host features, \textit{i.e.}, cavity size, functional groups, charge state, hydrophobicity, to the operating solvent regime, so the model is asked to reason from molecular structure rather than retrieve memorized study metadata.
The few-shot pool contains one example per solvent class, preferring rows where the guest also has a SMILES string and the host has not yet appeared in the pool; these examples are excluded from the evaluation set.

\subsection{Host-Guest Description}
\label{app:hgd}
The task has two complementary subtypes that share the same source.
For the \emph{forward} subtype, we group records by host, deduplicate guests by name, and keep only hosts that admit at least ten distinct guests with a known SMILES.
For each remaining host, we compute molecular-weight, hydrogen-bond-donor, hydrogen-bond-acceptor, ring-count, and formal-charge descriptors of every guest via RDKit, then take the top $30\%$ by $\log K_a$ as the high-affinity set.
The reference answer aggregates this set into a single paragraph that reports the mean molecular weight, the dominant formal charge, and the average counts of ring, H-bond donor, and H-bond acceptor sites, followed by a list of five representative guest names.
For the \emph{reverse} subtype, we group records by guest, drop guests with fewer than five distinct hosts, and again take the top $30\%$ by $\log K_a$.
The reference answer lists the top five hosts with their measured $\log K_a$ values, reports the maximum observed $\log K_a$, and summarizes the structural families of those hosts, \textit{e.g.}, cucurbituril, cyclodextrin, and calixarene, \textit{etc.}

\subsection{Molecular Identification}
\label{app:molecular_identification}

Based on the cleaned interaction data, we additionally construct a multimodal molecular identification benchmark using crawled two-dimensional molecular structure images from SupraBank. In this task, the model predicts the molecular name or canonical SMILES from a single molecular structure image. To improve evaluation robustness, we aggregate aliases from multiple molecular metadata sources, including common names, abbreviations, and IUPAC names.
Moreover, to ensure reliable alignment between molecular images and molecular annotations, all image records are linked via interaction-level identifiers rather than string-based name matching.
The final multimodal benchmark contains 1{,}773 unique molecular images and corresponding gold SMILES.

\begin{table*}[!t]
\centering
\small
\setlength{\tabcolsep}{5pt}
\renewcommand{\arraystretch}{1.05}
\resizebox{\linewidth}{!}{
\begin{tabular}{llcccccc}
\toprule
{} & \multirow{2}{*}{\textbf{Model}}
  & \multicolumn{6}{c}{\textbf{Molecular Identification}} \\
\cmidrule(lr){3-8}
& & Validity $\uparrow$ & Canonical $\uparrow$ & InChIKey $\uparrow$ & Formula $\uparrow$ & Tanimoto $\uparrow$ & $\Delta$Heavy $\downarrow$ \\
\midrule
\multirow{4}{*}{\rotatebox{90}{Base}}
   & Qwen3.5-9B     & $\underline{0.805}$ & $0.295$              & $0.390$              & $0.445$              & $0.696$              & $1.549$              \\
   & Qwen3.5-27B    & $0.738$              & $\underline{0.384}$ & $\underline{0.489}$ & $\underline{0.529}$ & $\underline{0.814}$ & $\mathbf{0.531}$    \\
   & GPT-5.4-Nano   & $0.648$              & $0.081$              & $0.099$              & $0.118$              & $0.385$              & $6.014$              \\
   & Gemini-3-Flash & $\mathbf{0.945}$    & $\mathbf{0.584}$    & $\mathbf{0.764}$    & $\mathbf{0.801}$    & $\mathbf{0.914}$    & $\underline{0.839}$ \\
\midrule
\multirow{4}{*}{\rotatebox{90}{Few-Shot}}
   & Qwen3.5-9B     & $0.832$              & $0.329$              & $0.436$              & $0.488$              & $0.722$              & $1.426$              \\
   & Qwen3.5-27B    & $\underline{0.894}$ & $\underline{0.450}$ & $\underline{0.576}$ & $\underline{0.620}$ & $\underline{0.798}$ & $\underline{0.685}$ \\
   & GPT-5.4-Nano   & $0.653$              & $0.106$              & $0.116$              & $0.140$              & $0.449$              & $4.076$              \\
   & Gemini-3-Flash & $\mathbf{0.958}$    & $\mathbf{0.593}$    & $\mathbf{0.770}$    & $\mathbf{0.800}$    & $\mathbf{0.915}$    & $\mathbf{0.369}$    \\
\midrule
\multirow{4}{*}{\rotatebox{90}{CoT}}
   & Qwen3.5-9B     & $\underline{0.799}$ & $0.294$              & $0.388$              & $0.438$              & $0.700$              & $4.271$              \\
   & Qwen3.5-27B    & $0.626$              & $\underline{0.380}$ & $\underline{0.455}$ & $\underline{0.507}$ & $\underline{0.838}$ & $\mathbf{0.378}$    \\
   & GPT-5.4-Nano   & $0.669$              & $0.084$              & $0.098$              & $0.115$              & $0.376$              & $6.325$              \\
   & Gemini-3-Flash & $\mathbf{0.955}$    & $\mathbf{0.567}$    & $\mathbf{0.755}$    & $\mathbf{0.800}$    & $\mathbf{0.904}$    & $\underline{0.526}$ \\
\bottomrule
\end{tabular}
}
\caption{Performance comparison for molecular identification task. The best score is shown in \textbf{bold}, and the second-best is \underline{\text{underlined}}.}
\label{tab:vision}
\end{table*}

\section{Text Corpus Construction Details}
\label{app:text_corpus}

\subsection{Text Collection}
\label{app:text_corpus_collection}

\paragraph{Anchor Corpus.}
We pull the Europe PMC abstract index through the public REST search endpoint, spanning publication years $1900$ through $2026$ and yielding approximately $40$ million abstract records.
In parallel, we mirror the bulk open-access full-text XML corpus, which covers approximately $8$ million articles, together with the auxiliary PMID, PMCID, and DOI identifier-mapping tables that link abstract metadata to full-text bodies.

\paragraph{Supramolecular Topical Queries.}
To obtain the domain-specific text corpus, we leverage $19$ topical Europe PMC search queries (shared by domain experts) that span the field's principal sub-areas, listed in Table~\ref{tab:corpus_keywords}.
We deduplicate the union of query results by PMID, yielding a raw split of $420{,}950$ unique articles.
Since Europe PMC matches against MeSH headings and indexed keyword lists in addition to titles and abstracts, the raw split retains residual biomedical contamination, such as articles that mention \emph{host cells} in immunology rather than \emph{host molecules} in supramolecular chemistry, which motivates us for the following two-stage rule-based filtering.

\paragraph{Two-Stage Rule-Based Filter.}
We refine the raw split with a transparent filter built from three keyword banks listed in Table~\ref{tab:corpus_keywords}.
The \emph{positive bank} of approximately $440$ supramolecular-specific keywords is organized into five tiers spanning core terms, named entities, and qualified self-assembly phrases.
The \emph{hard-reject bank} of approximately $140$ biomedical keywords in the title conclusively indicates off-topic content regardless of context.
The \emph{conditional-reject bank} of approximately $50$ keywords fires only when no supramolecular anchor is present in the title.
Given these banks, each article is deterministically assigned to one of five clusters based on where positive or negative signals appear in its title and abstract:
(A) title-positive, kept;
(B) abstract-positive with title-neutral, kept;
(C) hard-reject, dropped;
(D) conditional-reject, dropped;
(E) no signal in either field, excluded.
The two kept clusters together yield a filtered split of $133{,}867$ high-precision supramolecular articles.

\begin{table}[t]
\centering
\small
\setlength{\tabcolsep}{4pt}
\renewcommand{\arraystretch}{1.2}
\begin{tabular}{p{0.22\linewidth} p{0.68\linewidth}}
\toprule
\textbf{Bank} & \textbf{Representative keywords} \\
\midrule
\textbf{Topical queries} ($19$) & supramolecular, host–guest chemistry, self-assembly, molecular recognition, cyclodextrin, crown ether, calixarene, cucurbituril, pillararene, rotaxane/catenane, dendrimer, metal-organic framework, non-covalent interaction, molecular cage, inclusion complex, supramolecular polymer, metallosupramolecular, macrocyclic chemistry, binding affinity. \\
\midrule
\textbf{Positive bank} ($\sim$$440$) & supramolecular, host–guest, cucurbit[n]uril, calix[n]arene, pillar[n]arene, cryptand, cavitand, foldamer, mechanically interlocked, MOF, COF, $\beta$-CD, CB[7], MIL-53, MOF-5, HKUST-1, UiO-66, polyoxometalates, fullerenes, ionophores, $\ldots$ \\
\midrule
\textbf{Hard-reject bank} ($\sim$$140$) & SARS-CoV-2, T-cell receptor, bacterial infection, RNA splicing, viral capsid, immune response, $\ldots$ \\
\midrule
\textbf{Conditional-reject bank} ($\sim$$50$) & host cell, drug delivery, nanoparticle, liposome, protein-protein interaction, DNA binding, $\ldots$ \\
\bottomrule
\end{tabular}
\caption{Keyword banks used for the Europe PMC supramolecular corpus.}
\label{tab:corpus_keywords}
\end{table}

\subsection{Corpus Quality Validation}
\label{app:text_corpus_validation}

To quantify how much the two-stage filter actually concentrates supramolecular content, we employ an LLM-as-judge validation on randomly drawn samples from both splits.
Specifically, we randomly draw $5{,}000$ articles from the raw and filtered splits, present the title and abstract of each article to \texttt{Claude-Haiku-4.5}, and ask the model to classify the paper as relevant to the supramolecular chemistry domain on a strict centrality criterion: 
The paper is centrally about supramolecular chemistry only if its primary subject is non-covalent host–guest association, molecular recognition by macrocyclic hosts, non-covalent inclusion complexes, or supramolecular self-assembly. 
Table~\ref{tab:corpus_validation} reports the result.

\begin{table}[t]
\centering
\small
\resizebox{\linewidth}{!}{
\begin{tabular}{lcccc}
\toprule
\textbf{Split}  & \textbf{YES} $\uparrow$ & \textbf{BORDERLINE} & \textbf{NO} $\downarrow$ & \textbf{Positive} $\uparrow$ \\
\midrule
Raw              & $15.2\%$ & $3.1\%$ & $81.7\%$ & $18.3\%$ \\
Filtered         & $62.9\%$ & $10.9\%$ & $26.2\%$ & $73.8\%$ \\
\bottomrule
\end{tabular}
}
\caption{Corpus quality validation over $5{,}000$ randomly sampled articles via \texttt{Claude-Haiku-4.5}.}
\label{tab:corpus_validation}
\end{table}

\section{Details for Domain Adaptation}
\label{app:dapt_recipe}

We report below the single DAPT recipe used to produce the domain-adapted variants in Section~\ref{sec:domain_adapt} of the main paper.

\paragraph{Training Corpus.}
Existing studies~\citep{french1999catastrophic, li2024examining} have shown that DAPT on a narrow domain in isolation tends to degrade general-purpose competence.
To this end, we follow EvoLM~\citep{qi2025evolm} that construct an $80/15/5$ token-fraction mix, where $80\%$ of the tokens come from the filtered supramolecular split we discussed in Section~\ref{app:text_corpus_collection}, $15\%$ from FineWeb-Edu~\citep{lozhkov2024fineweb-edu} for general-domain replay, and $5\%$ from Tulu-3 SFT mixture~\citep{lambert2024tulu3} flattened into question-answer pairs.

\paragraph{Training Recipe.}
We employ LoRA for DAPT, with parameters rank 32, alpha 64 and dropout rate 0.05 to every transformer block.
Training runs for one epoch at a peak learning rate of $1\!\times\!10^{-5}$ with a cosine schedule and $5\%$ warmup in bf16, sharded with FSDP across four A100 GPUs.
The per-device batch size is $1$ with gradient accumulation $4$, giving an effective batch size of $16$ at a sequence length of $4096$.
We apply the identical recipe to both Qwen3.5-9B and Llama3.1-8B, holding the optimization budget fixed across bases so that downstream score differences in Table~\ref{tab:domain_adaptation} can be attributed to the base model rather than to the adaptation procedure.

\section{\modelname Examples}
\label{app:additional_examples}

This appendix presents one worked example per task, including the full prompt sent to the model and the model's response.
We use Gemini-3-Flash under the Base prompting strategy as the running model.

\begin{figure*}[t]
    \centering
    \begin{tcolorbox}[colback=gray!5, colframe=black!70, title=Binding Affinity Prediction Example]
    \textbf{Prompt:} Predict the binding affinity ($\log K_a$) between the host molecule $p$-Sulfonatocalix[4]arene and the guest molecule Gadolinium(III) chloride, SMILES: Cl[Gd](Cl)Cl in aqueous solution at standard conditions. Provide a single numerical value. Put your final answer between \texttt{<answer></answer>}.

    \textbf{Reference $\log K_a$:} $3.94$
    \tcblower
    \textbf{Gemini-3-Flash response:} To predict the binding affinity ($\log K_a$) between $p$-sulfonatocalix[4]arene ($sC4$) and gadolinium(III) chloride ($GdCl_3$), we consider the supramolecular interactions in aqueous environment. $p$-Sulfonatocalix[4]arene is a macrocyclic host with a hydrophobic cavity capped with four sulfonic acid groups at the upper rim and four phenolic hydroxyl groups at the lower rim; in aqueous solution near neutral pH, the sulfonic acid groups are deprotonated and give the host a charge of $-4$. Gadolinium(III) is a trivalent lanthanide cation, and binding between $sC4$ and lanthanide ions is primarily driven by strong electrostatic interactions between the negatively charged sulfonated rim and the highly charged $Gd^{3+}$ cation, supplemented by outer-sphere coordination to the sulfonic and phenolic oxygens. Experimental data for the binding of lanthanide(III) ions with $sC4$ typically show $\log K_a$ in the range of $3.5$ to $5.0$ depending on buffer and ionic strength.

    \texttt{<answer>4.15</answer>}
    \end{tcolorbox}
    \caption{Example for binding affinity prediction.}
    \label{fig:example_task1}
\end{figure*}

\begin{figure*}[t]
    \centering
    \begin{tcolorbox}[colback=gray!5, colframe=black!70, title=Top-Binder Selection Example]
    \textbf{Prompt:} Which of the following guest molecules has the strongest binding affinity (highest $\log K_a$) with the host molecule (1,1',4,4')-Tetraoxy-propylsulphite-bisnaftlen-methylen-bisglycouril Sodium in aqueous solution at standard conditions?
    \begin{itemize}\setlength{\itemsep}{0pt}
        \item[A.] Azure A
        \item[B.] N-Methylquinolinium iodide
        \item[C.] Methylene Blue
        \item[D.] sodium 3,3'-(1,3,6,8-tetraoxo-...-2,7-diyl)dipropionate
    \end{itemize}
    Put your final answer between \texttt{<answer></answer>}.

    \textbf{Reference:} \textbf{C} (Methylene Blue, $\log K_a = 7.60$). Distractor affinities: A $6.77$, B $3.90$, D $3.79$.
    \tcblower
    \textbf{Gemini-3-Flash response:} The host molecule, an acyclic naphthalene-walled cucurbituril analogue functionalized with sulfonate groups, possesses a hydrophobic cavity expanded by naphthalene walls and a strongly anionic rim that provides electrostatic complementarity to cationic guests. Azure A (A) and Methylene Blue (C) are cationic phenothiazine dyes that bind well to such hosts via combined hydrophobic and electrostatic interactions; N-methylquinolinium iodide (B) is a smaller cationic guest whose smaller surface area limits hydrophobic contact, and the sodium dipropionate-functionalized naphthalene diimide in (D) is an anion that experiences electrostatic repulsion. Methylene Blue in particular forms a rigid planar tricyclic system that fits perfectly between the naphthalene walls, yielding an exceptionally high $\log K_a$ exceeding that of the other three candidates.

    \texttt{<answer>C</answer>}
    \end{tcolorbox}
    \caption{Example for top-binder selection.}
    \label{fig:example_task2}
\end{figure*}

\begin{figure*}[t]
    \centering
    \begin{tcolorbox}[colback=gray!5, colframe=black!70, title=Host-Guest Description Forward Examp]
    \textbf{Prompt:} For the host molecule Cucurbit[8]uril, describe the key molecular properties of guest molecules that exhibit strong binding affinity (high $\log K_a$) in aqueous solution at standard conditions. Include: approximate molecular weight range, typical formal charge, number of ring systems, and any characteristic functional groups. Put your final answer between \texttt{<answer></answer>}.

    \textbf{Reference:} High-affinity guests of Cucurbit[8]uril ($\log K_a \geq 6.2$) typically have a molecular weight of $358 \pm 292$ g/mol, are neutral (formal charge $+0$), and contain on average $3.2$ ring systems, $2.3$ H-bond donors, and $3.2$ H-bond acceptors. Representative guests: L-phenylalanyl-L-valine, pentacyclo[7.3.1.1$^{4,12}$.0$^{2,7}$.0$^{6,11}$]tetradecane-4,9-diol, L-Tyrosyl-L-Phenylalanyl-L-alanine, 3,6-Diaminoacridin-10-ium, Tri-Trp Gly$_4$ target.
    \tcblower
    \textbf{Gemini-3-Flash response:} Cucurbit[8]uril (CB[8]) is a macrocyclic host composed of eight glycoluril units with a large hydrophobic cavity ($\sim$$479$ \AA$^3$) and two carbonyl-fringed portals. Strong binders fall into two regimes: monomeric guests with molecular weights between $200$ and $500$ Da that are large enough to maximize van der Waals contact with the cavity, and ternary complexes in which two smaller guests ($100$--$250$ Da each) jointly fill the cavity. The carbonyl portals provide strong ion-dipole stabilization, so high-affinity guests are almost always cationic (formal charge $+1$ to $+2$, occasionally higher for polyamines). Guests typically contain two to four fused aromatic rings, with viologen and methylviologen scaffolds, naphthalenediimide, and dicationic aromatic dyes (\textit{e.g.}, methylene blue, acridine) as archetypes. Characteristic functional groups include quaternary ammonium centers, aromatic $\pi$-rich cores, and amide carbonyls capable of hydrogen bonding to the portals.
    \end{tcolorbox}
    \caption{Example for the forward subtype of host–guest description.}
    \label{fig:example_task3_forward}
\end{figure*}

\begin{figure*}[t]
    \centering
    \begin{tcolorbox}[colback=gray!5, colframe=black!70, title=Host-Guest Description Reverse Example]
    \textbf{Prompt:} For the guest molecule 2-phenylethylamine hydrochloride (SMILES: NCCc1ccccc1.[H]Cl), describe the key molecular properties of host molecules that exhibit strong binding affinity (high $\log K_a$) in aqueous solution at standard conditions. Include: host structural type, approximate cavity size or geometry, typical charge, and any characteristic binding interactions. Put your final answer between \texttt{<answer></answer>}.

    \textbf{Reference:} High-affinity hosts for 2-phenylethylamine hydrochloride ($\log K_a \geq 1.9$) include $5{,}11{,}17{,}23$-tetrakis(dihydroxyphosphoryl)-$25{,}26{,}27{,}28$-tetrapropoxycalix[4]arene ($\log K_a = 2.2$) and a sulfonatobutoxy-substituted tris-oxazoline triethylbenzene host ($\log K_a = 1.9$). The highest recorded $\log K_a$ is $2.2$. Favorable hosts are predominantly calixarene macrocycles (cone-shaped, phenol-derived).
    \tcblower
    \textbf{Gemini-3-Flash response:} 2-Phenylethylamine hydrochloride is amphiphilic, with a hydrophobic phenyl ring and a hydrophilic, positively charged $-$NH$_3^+$ group. Effective hosts must accommodate both moieties. The preeminent candidate is Cucurbit[7]uril (CB[7]), a macrocycle of seven glycoluril units forming a barrel-shaped cavity ($\sim$$279$ \AA$^3$, $\sim$$5$--$7$ \AA diameter, $\sim$$9.1$ \AA height); the hydrophobic cavity encapsulates the phenyl ring with high packing density, the carbonyl-lined portals stabilize the ammonium by ion-dipole interactions, and CB[7]-arylalkylammonium complexes typically reach $\log K_a$ between $5$ and $7$. Cyclodextrins (especially $\beta$-CD) and sulfonated calixarenes are weaker but viable alternatives: $\beta$-CD provides a $\sim$$6.0$ \AA cavity for hydrophobic inclusion of the phenyl ring, and $p$-sulfonatocalix[4]arene supplements the hydrophobic effect with cation-$\pi$ interactions and electrostatic stabilization of the ammonium.
    \end{tcolorbox}
    \caption{Example for the reverse subtype of host–guest description.}
    \label{fig:example_task3_reverse}
\end{figure*}

\begin{figure*}[t]
    \centering
    \begin{tcolorbox}[colback=gray!5, colframe=black!70, title=Solvent Identification Example]
    \textbf{Prompt:} Given a host–guest complex, predict which solvent environment is most appropriate for measuring its binding constant. The host molecule is provided as a SMILES string; analyze its structure (cavity size, functional groups, charge, hydrophobicity) to determine the solvent class. The candidate solvents are 
    \begin{itemize}\setlength{\itemsep}{0pt}
        \item[A.] Water
        \item[B.] DMSO
        \item[C.] MeCN
        \item[D.] MeOH
        \item[E.] CHCl$_3$
        \item[F.] CH$_2$Cl$_2$
    \end{itemize}
    Host SMILES: \textit{[skipped for simplicity]}. Put your final answer between \texttt{<answer></answer>}.
    \textbf{Reference:} \textbf{A} (water).
    \tcblower
    \textbf{Gemini-3-Flash response:} The provided SMILES represents $p$-sulfonatocalix[4]arene. Its structure is a calix[4]arene scaffold of four phenolic rings linked by methylene bridges that defines a hydrophobic cavity; the upper rim is substituted with four sulfonic acid groups ($-$SO$_3$H) that are typically deprotonated to sulfonates ($-$SO$_3^-$) in solution. The four sulfonate groups make the molecule highly anionic and extremely soluble in polar, aqueous environments. $p$-Sulfonatocalix[4]arenes are archetypal water-soluble hosts: they drive guest inclusion through the hydrophobic effect into the cavity, while the sulfonate rim provides water solubility and electrostatic stabilization of cationic guests. By the operating-medium rule for sulfonated macrocycles with polar portals, the appropriate solvent is water.

    \texttt{<answer>A. water</answer>}
    \end{tcolorbox}
    \caption{Example for solvent identification.}
    \label{fig:example_task7}
\end{figure*}

We further include in Figure~\ref{fig:case_study_cot} the full Base and Chain-of-Thought traces behind the failure case analyzed in Section~\ref{sec:exp:case_study}, where Chain-of-Thought reasoning destabilizes an otherwise near-perfect binding affinity prediction.
\begin{figure*}[t]
    \centering
    \begin{tcolorbox}[colback=gray!5, colframe=black!70, title=Chain-of-Thought Failure on Binding Affinity Prediction (DeepSeek-v4)]
    \textbf{Prompt:} Predict the binding affinity ($\log K_a$) between the host molecule 4,7,13,18-Tetraoxa-1,10-diazabicyclo[8.5.5]icosane (SMILES: \texttt{C1COCCN2CCOCCN1CCOCCOCC2}) and the guest molecule Barium cation (SMILES: \texttt{[Ba+2]}) in aqueous solution at standard conditions. Provide a single numerical value. Put your final answer between \texttt{<answer></answer>}.

    \textbf{Reference $\log K_a$:} $2.00$
    \tcblower
    \textbf{Base response:} \texttt{<answer>2.1</answer>} \quad\textcolor{teal}{\textbf{(correct)}}

    \vspace{0.6em}
    \textbf{CoT response}: The host molecule is a diaza-crown ether derivative, specifically a bicyclic structure with two nitrogen atoms and four ether oxygen atoms. This type of ligand is known to bind strongly to alkali and alkaline earth metal cations, particularly Ba$^{2+}$, due to its preorganized cavity and favorable ion--dipole interactions. In aqueous solution, the binding constant for Ba$^{2+}$ with such bicyclic diaza-crown ethers typically falls in the range of $\log K_a \approx 10$--$12$, depending on ring size and solvation effects. Given the structure (a [8.5.5] system), it is a relatively good fit for Ba$^{2+}$. Considering standard aqueous conditions and typical literature values for similar cryptands, a reasonable estimate is:

    \texttt{<answer>11.0</answer>} \quad\textcolor{red}{\textbf{(off by nine orders of magnitude in $K_a$)}}
    \end{tcolorbox}
    \caption{Case study of a Chain-of-Thought regression. On the same host--guest pair, the Base prompt returns a near-perfect $\log K_a$, while CoT articulates fluent but uncalibrated chemistry and fabricates a ``typical literature'' range that does not exist, landing nine orders of magnitude away in the association constant. See Section~\ref{sec:exp:case_study}.}
    \label{fig:case_study_cot}
\end{figure*}

\end{document}